%% file: agent_cybernetics_arxiv.tex
\documentclass{article}

\PassOptionsToPackage{sort&compress}{natbib}

\usepackage[preprint]{neurips_2026}

\usepackage[utf8]{inputenc}
\usepackage[T1]{fontenc}
\usepackage{hyperref}
\usepackage{url}
\usepackage{booktabs}
\usepackage{amsfonts}
\usepackage{nicefrac}
\usepackage{microtype}
\usepackage{xcolor}
\usepackage{enumitem}
\usepackage{amsmath,amssymb,amsthm}
\usepackage{multirow}
\usepackage{makecell}

\newtheorem{theorem}{Theorem}
\newtheorem{principle}{Principle}

\newtheorem{desideratum}{Desideratum}

\usepackage{tcolorbox}
\tcbuselibrary{skins,breakable}

\newtcolorbox{desideratabox}[2][]{
  enhanced, colback=blue!4!white, colframe=blue!50!black,
  fonttitle=\bfseries, title={#2}, #1
}
\newtcolorbox{engineeringbox}[2][]{
  enhanced, colback=green!4!white, colframe=green!50!black,
  fonttitle=\bfseries, title={#2}, #1
}
\newtcolorbox{failurebox}[2][]{
  enhanced, breakable, colback=red!4!white, colframe=red!40!black,
  fonttitle=\bfseries\small, title={#2},
  left=6pt, right=6pt, top=4pt, bottom=4pt, #1
}
\newtcolorbox{insightbox}[2][]{
  enhanced, breakable, colback=blue!4!white, colframe=blue!45!black,
  fonttitle=\bfseries\small, title={#2},
  left=6pt, right=6pt, top=4pt, bottom=4pt, #1
}
\newtcolorbox{recbox}[2][]{
  enhanced, breakable, colback=green!4!white, colframe=green!45!black,
  fonttitle=\bfseries\small, title={#2},
  left=6pt, right=6pt, top=4pt, bottom=4pt, #1
}

\title{The Agent Use of Agent Beings: 
Agent Cybernetics Is the Missing Science of Foundation Agents}

\author{%
Xinrun Wang$^1$\thanks{Equal Contribution. Email: xrwang@smu.edu.sg, chang.yang@connect.polyu.hk}~, Chang Yang$^2$\footnotemark[1]~, He Zhao$^3$, Zhuoyi Lin$^4$, Shuyue Hu$^5$ \\
$^1$Singapore Management University~~$^2$The Hong Kong Polytechnic University \\
$^3$Nanyang Technological University\\ $^4$Institute for Infocomm Research, Agency for Science, Technology and Research (A*STAR) \\
$^5$Shanghai Artificial Intelligence Laboratory
}

\begin{document}

\maketitle

\begin{abstract}
LLM-based foundation agents that perceive, reason, and act across thousands of reasoning steps are rapidly becoming the dominant paradigm for deploying artificial intelligence in open-ended, long-horizon complex tasks. Despite this significance, the field remains overwhelmingly engineering-driven. Engineering practice has converged on useful primitives (tool loops, memory banks, harnesses, reflection steps), yet these are assembled by empirical trial and error rather than from first principles. Fundamental questions remain open: under what conditions does a long-running agent remain on-task? How should an agent respond when its environment exceeds its representational capacity? What architectural properties are necessary for safe self-improvement?
We argue that \emph{cybernetics}, the mid-twentieth-century science of control and communication in complex systems, provides the missing theoretical scaffold for foundation agents. By mapping six canonical laws of classical cybernetics onto six agent design principles, and synthesizing those principles into three engineering desiderata (reliability, lifelong running, and self-Improvement), we arrive at a framework termed \textbf{Agent Cybernetics}. Three application domains, code generation, computer use and automated research, exemplify the analytical framework of agent cybernetics by identifying failure modes and concrete engineering recommendations. We hope that agent cybernetics opens a new research venue and establishes 
the scientific foundation that foundation agents need for principled, reliable real-world deployment.
\end{abstract}

\vspace{-15pt}
\section{Introduction}
\vspace{-5pt}
Norbert Wiener's \emph{The Human Use of Human Beings}~\citep{wiener1988human} argued that the central challenge of the coming machine age would not be building capable components, but designing systems in which those components interact reliably under uncertainty. That challenge has arrived with the rapid development of foundation agents, where a large pre-trained language model (LLM) serves as the cognitive core of an autonomous loop that perceives, reasons, acts, and revises its own behavior over extended horizons. Unlike single-shot inference, such agents accumulate state, invoke external tools, and operate across thousands of reasoning steps in pursuit of complex, open-ended goals. The celebrating progress of foundation agents spans across software engineering~\citep{jimenez2024swebench,deng2025swe}, computer use~\citep{xie2024osworld,bonatti2024windows}, math proof~\citep{trinh2024solving,hubert2025olympiad}, and scientific discovery~\citep{zhang2025multimodal,gao2026autonomous,feng2026internagent}. 
Engineering practice has converged on useful primitives: tool loops~\citep{qin2024tool,xue2025simpletir}, memory banks~\citep{chhikara2025mem0,sumers2023cognitive}, harnesses~\citep{zhou2026externalization,lee2026meta}, and reflection steps~\citep{shinn2023reflexion,tan2025cradle}. 
However, these successes are assembled by empirical trial and error rather than by principled design. Questions that any mature engineering discipline would consider fundamental remain open: Under what conditions does a long-running agent remain on-task? How should an agent respond when its environment exceeds its representational capacity? What architectural properties are necessary for an agent to improve its own performance? In summary, the fundamental question to the researchers is:
\begin{tcolorbox}[
  left=4pt, right=4pt, top=5pt, bottom=5pt,
  boxsep=0pt
]
\centering
\emph{Can we build the \textbf{science} of foundation agents? If yes, what should it look like?}
\end{tcolorbox}

\emph{Cybernetics}, i.e., the science of control and communication in complex systems, emerged in the mid-twentieth century as a unified framework for understanding purposive behavior under uncertainty. Its foundational contributions span Wiener's mathematical theory of feedback and control~\citep{wiener1949cybernetics}, Ashby's formalization of adaptive regulation and requisite variety~\citep{ashby1956introduction,ashby1960design}, and Shannon's information-theoretic account of reliable communication over noisy channels~\citep{shannon1948mathematical}. Von Foerster subsequently extended this programme into second-order cybernetics, introducing self-reference and the observer into the system's own description~\citep{von2003cybernetics}. A particularly prescient instantiation is Qian Xuesen's \emph{Engineering Cybernetics}~\citep{qian1954engineering}, which addressed the central engineering challenge of constructing reliable, goal-directed systems from inherently unreliable components.

\textbf{Our Position.} We argue that \textbf{cybernetics provides the principled scaffold of foundation agents}. Modern language models are precisely the kind of unreliable modules Qian envisioned: prone to hallucination under distribution shift, goal drift over long contexts, and silent failure on edge cases. \emph{Agent Cybernetics} maps six canonical laws of classical cybernetics onto six agent design principles, and synthesizes those principles into three engineering desiderata (i.e., reliability, lifelong running, and self-improvement) whose derivation is developed in Sections~\ref{sec:classical}-\ref{sec:agent}. Section~\ref{sec:suggested_research} translates these desiderata into a concrete research agenda. We hope Agent Cybernetics can establish the scientific principles for foundation agents towards real-world deployment.

% \textbf{Our Position.} We argue that \emph{cybernetics}, the science of control and communication in complex systems, independently co-discovered by Wiener~\citep{wiener1949cybernetics}, Ashby~\citep{ashby1956introduction,ashby1960design}, and Shannon~\citep{shannon1948mathematical}, and extended into second-order self-reference by von Foerster~\citep{von2003cybernetics}, provides the missing principled scaffold. Cybernetics was itself motivated by the challenge of engineering reliable purposive behavior in systems built from unreliable components operating under uncertainty. That description applies precisely to today's LLM-based agents. Xuesen Qian's foundational programme of \emph{Engineering Cybernetics}, i.e., building reliable systems from unreliable modules, is particularly prescient: a modern language model is precisely such an unreliable module, prone to hallucination under distribution shift, goal drift under long contexts, and silent failure on edge cases~\citep{qian1954engineering}.

\begin{figure}[t]
    \centering
    \vspace{-30pt}
    \includegraphics[width=\linewidth]{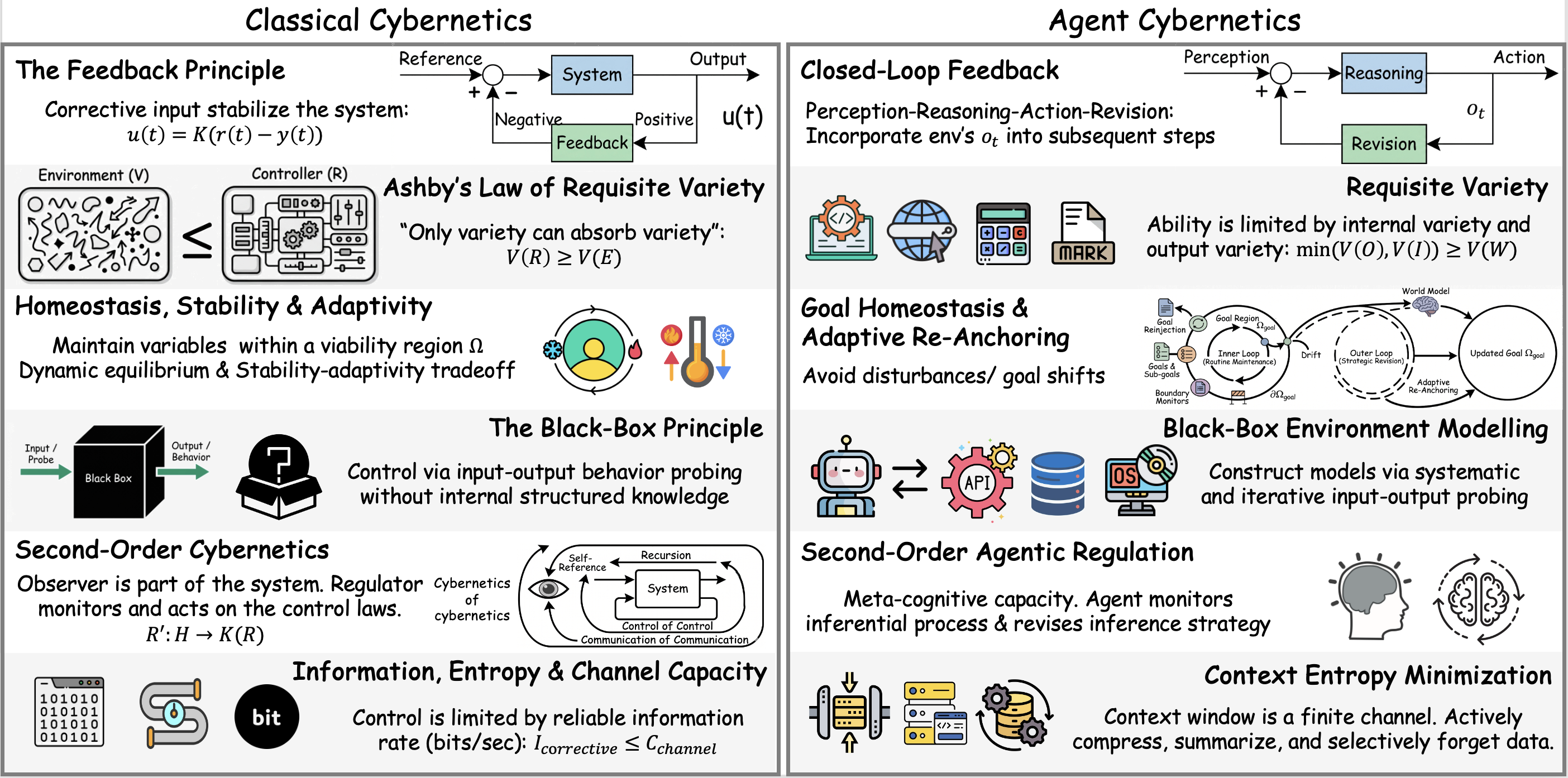}
    \caption{From Classical Cybernetics to Agent cybernetics}
    \label{fig:ac}
    \vspace{-10pt}
\end{figure}

\vspace{-5pt}
\section{Classical Cybernetics: Core Laws}
\label{sec:classical}

Cybernetics emerged in the 1940s as a unified theory of purposive, self-regulating systems, synthesizing control engineering, neurophysiology, statistical mechanics, and information theory. We survey its six most consequential formal results as the foundation for \textbf{Agent Cybernetics} in Section~\ref{sec:agent}.

\textbf{The Feedback Principle.}
A system $E$ is \emph{closed-loop} if its output $y(t)$ is routed back as a corrective input $u(t)$ to the regulator~\citep{wiener1949cybernetics}:
$u(t) = K\bigl(r(t) - y(t)\bigr),$
where $r(t)$ is the reference signal and $K$ is the controller gain. \emph{Negative feedback} suppresses deviations from $r$, conferring robustness to perturbations, while \emph{positive feedback} amplifies deviations, driving the system toward a new equilibrium or instability. An open-loop system ($u$ independent of $y$) cannot self-correct: perturbations accumulate without bound unless the plant is intrinsically stable.

\textbf{Ashby's Law of Requisite Variety.}
Let $V(E)$ denote the \emph{variety} of a system $E$, i.e., the logarithm of the number of distinguishable states, and let $V(R)$ denote the variety of its regulator $R$~\citep{ashby1956introduction}. Only variety can absorb variety (Theorem~\ref{theorem:variety}): a regulator whose state space is smaller than that of its environment cannot neutralize all disturbances. This result has a consequential implication for agent design: capability is ultimately a function of state-space coverage, not only computational power.

\begin{theorem}[Law of Requisite Variety]
\label{theorem:variety}
Complete control of $E$ by $R$ requires $V(R) \geq V(E)$. The residual variety in the regulated output satisfies $V(E') \geq V(E) - V(R)$.
\end{theorem}

\textbf{Homeostasis, Stability, and Adaptivity.}
Cannon's physiological notion of homeostasis~\citep{cannon1939wisdom}, formalized by Ashby~\citep{ashby1960design}, describes the capacity of a system to maintain its essential variables within a viability region $\Omega \subset \mathcal{X}$ despite external perturbations:
$x(t) \in \Omega,\forall t \geq 0,$
where $x(t)$ is the system state. Homeostatic regulation activates corrective actions whenever $x(t)$ approaches $\partial\Omega$. Crucially, homeostasis is \emph{dynamic equilibrium}: the system may traverse a wide region of $\mathcal{X}$ while continuously returning toward $\Omega$, and the boundary $\partial\Omega$ may itself adapt over time.
This dynamic character gives rise to a fundamental \emph{stability-adaptivity tradeoff}. Stability demands that coordination, internal representations, and goal-directed behavior not be destabilized by every novel stimulus. Adaptivity demands that rigid adherence to a fixed $\Omega$ not render the agent brittle under distributional shift or genuinely novel tasks.
Ashby's notion of \emph{ultrastability}~\citep{ashby1960design} offers a principled resolution through a two-level feedback architecture. A fast \emph{inner loop} preserves $x(t) \in \Omega$ under routine perturbations, exploiting current knowledge for reliable performance. A slow \emph{outer loop} restructures $\Omega$ itself when sustained boundary violations signal that the current viability region is inadequate, triggering principled revision rather than brittle failure. In this hierarchy, stability and adaptivity are not opposing forces but complementary operating modes at different timescales.

\textbf{The Black-Box Principle.}
Ashby~\citep{ashby1956introduction} and Wiener~\citep{wiener1949cybernetics} independently established that the internal constitution of a complex system is often inaccessible or irrelevant. Sufficient understanding for control can be obtained by systematically probing input-output (I/O) behaviors of the system being controlled. Two corollaries follow: behavior modeling does not presuppose structural transparency, and any sufficiently rich I/O history can substitute for structural knowledge, provided the probing strategy is designed to distinguish the system's responses.

\textbf{Second-Order Cybernetics.}
Von Foerster~\citep{von2003cybernetics} extended classical cybernetics by recognizing that observers are not external to the systems they study. \emph{Second-order cybernetics}, i.e., the cybernetics of cybernetics, holds that the observer's cognitive structure recursively shapes the observed system, making the boundary between system and regulator permeable. This self-referential view foregrounds \emph{self-organization}: systems can modify their own control laws. Formally, a second-order regulator $R'$ acts on the space of first-order regulators:
$R' : \mathcal{H} \rightarrow \mathcal{K}(R),$
where $\mathcal{H}$ is the performance history of $R$ and $\mathcal{K}(R)$ is the space of available control laws.

\textbf{Information, Entropy, and Channel Capacity.}
Wiener~\citep{wiener1949cybernetics} and Shannon~\citep{shannon1948mathematical} independently arrived at a foundational insight: control is ultimately an information-flow problem. Shannon's channel capacity theorem establishes that the reliable information rate of a noisy channel is bounded by its capacity $C$ (bits/s). Translated into the language of regulation, this yields the constraint
$
    I_{\text{corrective}} \leq C_{\text{channel}},
$
where $I_{\text{corrective}}$ denotes the information rate required to suppress disturbances. Any regulator that cannot acquire and transmit corrective signals at this rate will fail to regulate, which is not attributable to a suboptimal control law or insufficient actuator authority, but rather to the information channel itself becoming the bottleneck. The engineering implication is immediate: before optimizing the control policy, one must first ensure that the communication link is adequate.
An equivalent and dual statement follows directly: the entropy $H$ of the controlled output satisfies
$
    H(\text{output}) \geq H(E) - C_{\text{channel}},
$
establishing a hard lower bound on residual uncertainty. Even a perfect regulator cannot drive the output entropy to zero; whatever portion of the disturbance entropy $H(E)$ exceeds the channel capacity $C_{\text{channel}}$ is necessarily reflected in the output.

% Wiener~\citep{wiener1949cybernetics} and Shannon~\citep{shannon1948mathematical} converged independently on the insight that control is fundamentally an information-flow problem. Shannon's channel capacity theorem bounds the reliable information rate of a noisy channel with capacity $C$ (bits/s). Applied to regulation:
% $I_{\text{corrective}} \leq C_{\text{channel}},$
% where $I_{\text{corrective}}$ is the information rate required to suppress disturbances. A regulator unable to acquire and transmit corrective signals at rate $I_{\text{corrective}}$ necessarily fails to regulate. Equivalently, the entropy $H$ of the controlled output is lower-bounded by $H(E) - C_{\text{channel}}$.

\section{From Classical Cybernetics to Agent Cybernetics}
\label{sec:agent}

Foundation agents are \emph{discrete, high-dimensional, self-referential control systems}. The agent maintains a running context $c_t$ (the current token sequence), selects actions $a_t$ (tool calls, text generation, memory writes), and receives observations $o_t$ (tool results, environment feedback):
\begin{equation}
  c_{t+1} = f_\theta\!\left(c_t, a_t, o_t\right), \qquad a_t \sim \pi_\theta(\cdot \mid c_t),
  \label{eq:agent_dynamics}
\end{equation}
where $f_\theta$ is the context update rule and $\pi_\theta$ is the LLM-induced policy. This dynamical description makes the cybernetic correspondence precise: $c_t$ is the system state, $\pi_\theta$ is the regulator, and $o_t$ is the feedback signal. We now derive six agent principles corresponding to the classical laws of Section~\ref{sec:classical}.

\begin{figure}[t]
    \centering
    \vspace{-30pt}
    \includegraphics[width=\linewidth]{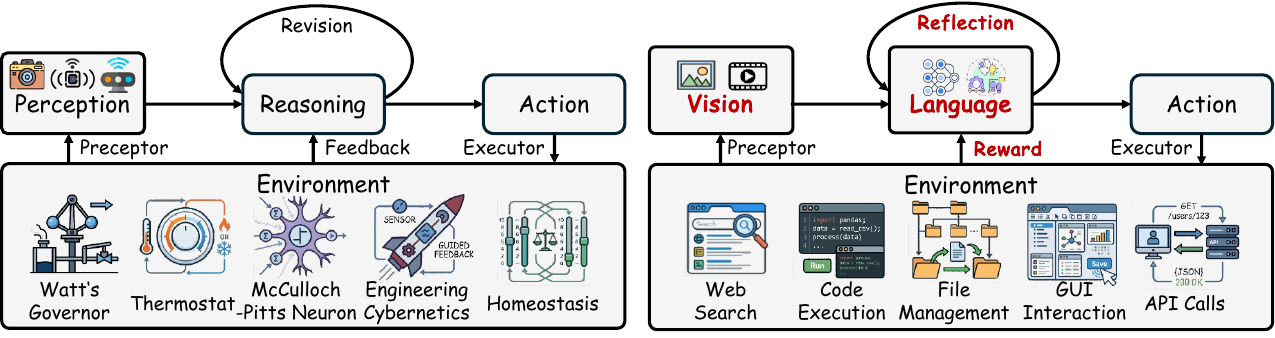}
    \caption{The terminology between classical cybernetics (left) and agent cybernetics (right). The terms with different names are highlighted as red. }
    \label{fig:classical_to_agent}
    \vspace{-10pt}
\end{figure}

\subsection{Principles in Agent Cybernetics}
\begin{principle}[Closed-Loop Feedback]
\label{prin:feedback}
The \textbf{perception-reasoning-action-revision} cycle of a foundation agent is the agentic analogue of the feedback principle. Effective agent design requires that environmental returns be faithfully incorporated into subsequent inference steps.
\end{principle}

Each execution loop of foundation agents constitutes an action $a_t$ and the returned result $o_t$ from the external environment. The agent's subsequent reasoning step is the regulatory response. A critical structural difference from continuous control is that feedback takes the form of a \emph{discrete, structured token sequence} rather than a real-valued measurement, such as the compiler errors and the search results. This makes \emph{feedback blindness}, i.e., the tendency of LLMs to regenerate prior plans without incorporating tool results, a particularly acute failure mode. Closed-loop design therefore requires explicit architectural mechanisms, such as structured prompts that mandate acknowledgment of the previous tool result before planning the next action.

% The Vision-Language-Action (VLA) paradigm~\citep{zitkovich2023rt2} instantiates the action-observation-revision cycle in embodied settings, yet current systems still lack robust revision capabilities~\citep{intelligence2026pi07steerable}. The missing \emph{revision} component, precisely the closed-loop feedback in the cybernetic sense, is a key architectural bottleneck limiting adaptability to novel tasks. The emerging literature on \emph{agent harnesses}~\citep{lou2026autoharness} addresses this concern at the engineering level: A harness enforces closed-loop execution by intercepting tool calls, structuring results, and re-injecting them into context in a form the model must act on. The cybernetic framing explains precisely why harnesses improve performance: they are feedback-closure mechanisms.

Vision-Language-Action (VLA)~\citep{zitkovich2023rt2} instantiates the action-observation-revision cycle in embodied settings. Nevertheless, existing systems exhibit significant deficiencies in their revision capabilities~\citep{intelligence2026pi07steerable}. The absence of robust closed-loop feedback constitutes a fundamental architectural bottleneck, systematically constraining adaptability to out-of-distribution tasks and novel operating environments. The emerging literature on \textbf{\emph{agent harnesses}}~\citep{lou2026autoharness} addresses this limitation at the engineering level: A harness enforces closed-loop execution by intercepting tool calls, structuring intermediate results, and re-injecting them into the model's context in a form that necessitates subsequent deliberation and action. From the cybernetics perspective, the performance gains attributable to harnesses demonstrates that the feedback-closure implementation is critical for stable performance.

\begin{principle}[Requisite Variety]
\label{prin:variety}
Let $V(\mathcal{O})$ denote the variety of an agent's \emph{output space}, i.e., the set of distinguishable responses, decisions, and action sequences the agent can produce, and $V(\mathcal{I})$ denote the variety of its \emph{internal representational space}, i.e., the set of distinguishable internal states the agent can occupy. Robust task completion against an environment of variety $V(\mathcal{W})$ requires:
$
  \min\bigl(V(\mathcal{O}),\, V(\mathcal{I})\bigr) \geq V(\mathcal{W}).
$
The binding constraint is whichever is smaller: an agent with a rich output space but impoverished internal representations cannot select the right output; an agent with rich representations but a constrained output space cannot externalize the appropriate response.
\end{principle}

The Law of Requisite Variety constrains agent design along two orthogonal dimensions. The \emph{output variety} $V(\mathcal{O})$ denotes the full space of actions available to the agent (i.e., tool invocations, generated text, memory writes, escalation decisions) and bounds the set of environmental states to which the agent can meaningfully respond. The \emph{internal variety} $V(\mathcal{I})$ denotes the representational capacity of the agent's context, world model, and memory structures, and bounds the set of environmental states the agent can \emph{distinguish}. Both dimensions are necessary conditions for effective control: environmental distinctions that cannot be internally represented will propagate as undifferentiated outputs, regardless of how rich the output space can be.

Tool sets constitute one mechanism for expanding $V(\mathcal{O})$, but the effective output variety depends not only on the cardinality of available tools but also on the agent's capacity for \emph{composition} and \emph{sequencing} them. A compact tool set with flexible compositional structure may realize greater effective output variety than a larger but rigidly invoked one. Analogously, $V(\mathcal{I})$ is expanded by richer context representations, more expressive world models, and hierarchical memory architectures that enable the agent to maintain fine-grained distinctions among environmental states.

Scaling model capacity can potentially increase both $V(\mathcal{I})$, through richer internal representations, and $V(\mathcal{O})$, through more expressive generation. However, such scaling can not fully resolve variety deficits arising from architectural constraints on context structure or action space. Ashby's Law further implies a \emph{selection cost}: an agent endowed with very large $V(\mathcal{O})$ and $V(\mathcal{I})$ must itself expend representational capacity to navigate its own state and output spaces. This observation motivates \textbf{hierarchical organization} as a design principle: both internal representations and output actions should be structured so that the agent operates within a compressed, task-relevant subspace, invoking the full variety space only when the compressed subspace proves insufficient.

\begin{principle}[Goal Homeostasis and Adaptive Re-Anchoring]
\label{prin:homeostasis}
Long-horizon agents are susceptible to \emph{goal drift}: the original task specification becomes semantically diluted over extended contexts, causing the effective objective to diverge from the intended one. Reliable agents require a two-level homeostatic architecture: i) a \textbf{fast inner loop} that maintains the current goal within a stability region $\Omega_{\text{goal}}$, and ii) a \textbf{slow outer loop} that restructures $\Omega_{\text{goal}}$ itself when sustained boundary violations signal that the current goal representation is inadequate. In multi-agent settings, individual agents may transiently leave their local goal viability regions in service of a shared system-level objective, requiring coordination mechanisms that balance local stability against collective adaptivity.
\end{principle}

Goal drift arises as a direct analogue of homeostatic failure. As the context $c_t$ accumulates intermediate observations, sub-task outputs, and error messages, the original task specification is progressively marginalized. The agent's implicit objective shifts toward satisfying local consistency conditions rather than the global task goal.
The two-level architecture addresses this failure mode by separating the timescales and mechanisms of goal maintenance from those of goal revision.
The inner loop counteracts this tendency through three complementary mechanisms: (i) periodic re-injection of the original task specification into the active context; (ii) explicit, maintained representations of current goals and sub-goals; and (iii) boundary monitors that detect when the agent's trajectory approaches $\partial\Omega_{\text{goal}}$ and trigger re-planning accordingly.
The outer loop governs \emph{adaptive re-anchoring}. When the original goal specification is persistently violated despite inner-loop corrections, indicating  either a fundamental change in the task or an inadequacy in the goal representation itself, the outer loop initiates goal revision. This may take the form of reformulating the goal, decomposing it into updated sub-goals, or escalating to a human overseer for clarification. The separation between the two levels is architecturally significant: conflating routine goal maintenance with strategic goal revision leads either to excessive rigidity or to uncontrolled objective drift.
World models serve as natural goal-anchoring structures within this framework. A world model provides a compact, defeasible representation of task-relevant state against which the agent continuously verifies its trajectory relative to $\Omega_{\text{goal}}$. It also constitutes the substrate for simulating candidate goal revisions before the outer loop commits to them, enabling anticipatory re-anchoring rather than purely reactive correction.

\begin{principle}[Black-Box Environment Modelling]
\label{prin:black_box}
An agent interacting with external APIs, databases, or operating systems typically lacks structural access to the environment. All prior knowledge should be treated as falsifiable hypotheses to be confirmed by observations rather than ground truths. Agents should construct working environment models through systematic, iterative input-output probing.
\end{principle}

The black-box principle reframes environmental opacity not as a deficiency to be overcome, but as the \textbf{expected epistemic condition} of any agent operating in an open-world setting. In practice, this means that an agent can rarely assume full knowledge of the system it interacts with: internal structure may be hidden, documentation may be incomplete, and the environment itself may change over time. This reframing carries a consequential design implication: confidence in prior structural knowledge should be calibrated to empirical track record rather than provenance. Documentation becomes stale, environments drift, and undocumented behaviors accumulate silently, which brings the silent divergence between an agent's internal model and the world it acts upon. 

Therefore, an agent that treats prior knowledge as ground truth is brittle by construction as when its internal model diverges from reality (inevitable in dynamic environments), the agent has no mechanism to detect and address the mismatch. An agent designed around the black-box principle behaves differently: instead of committing immediately to consequential actions, it first issues low-cost exploratory actions to probe the environment and update its beliefs. Rather than treating errors as terminal failures, it interprets them as informative observations that narrow down the space of plausible system behaviors. Furthermore, it will revise its actions incrementally as new evidence arrives to handle such changes. This distinction between \emph{assumption-driven} and \emph{evidence-driven} operation is not merely philosophical but practical: a reliable agent should not assume the environmental stability, but adapt to achieve the stability in the unstable environments.

\begin{principle}[Second-Order Agent Regulation]
\label{prin:second_order}
A sufficiently autonomous agent must monitor and regulate its own inferential process, not merely its external actions. This meta-cognitive capacity encompasses detecting looping behavior, declining confidence, and reasoning inconsistencies, and responding by revising the inference strategy rather than the immediate action.
\end{principle}

Von Foerster's central insight was that the observer cannot be cleanly separated from the observed system: the cognitive structure of the regulator recursively shapes what it can perceive and therefore what it can control~\cite{von2003cybernetics}. For foundation agents, this principle demarcates a qualitative boundary between two distinct levels of regulation.
A \emph{first-order} agent selects actions conditioned solely on environmental state; consequently, its failure modes are environmental in origin and can in principle be corrected by environmental feedback. A \emph{second-order} agent additionally monitors the \emph{structure} of its own reasoning process, i.e., self-improving~\citep{zhang2026hyperagents,zhang2025darwin}. Concretely, this includes detecting that the agent has issued structurally identical actions across many consecutive steps (indicative of a reasoning loop), that its confidence estimates have been systematically mis-calibrated relative to realized outcomes, or that its chain of inference has drifted from the original task specification. These are failures that no  environmental feedback can correct, because they originate in the regulatory process itself rather than in the environment being regulated.
Let $R$ denote a first-order regulator operating over an environmental state space. A second-order regulator $R'$ acts not on the environment directly, but on the space of first-order regulators themselves:
$
    R' : \mathcal{H} \rightarrow \mathcal{K}(R),
$
where $\mathcal{H}$ denotes the performance history of $R$ (e.g., action sequences, trajectories, performance scores), and $\mathcal{K}(R)$ denotes the space of available control laws that can be substituted for or applied to $R$. In operational terms, this formulation encompasses self-consistency verification, loop detection, confidence-gated escalation to a human or supervisory agent, and dynamic revision of inference strategy.

\begin{principle}[Context Entropy Minimization]
\label{prin:memory}
The context window of an LLM is a finite-capacity information channel. Appending low-information content increases context entropy without reducing task uncertainty, thereby degrading effective channel capacity. Agentic architectures should incorporate active context compression and selective memory that retains information if and only if its expected contribution to reducing future decision uncertainty exceeds a threshold.
\end{principle}

Every token appended to the context window consumes capacity that could carry task-relevant signal. Without explicit architectural intervention, long-horizon agent contexts accumulate indiscriminately: verbose tool outputs, superseded plans, redundant intermediate steps, and failed attempts accrete without bound, progressively diluting the signal that governs correct action selection. The Shannon--Wiener channel capacity principle makes this cost precise: the entropy $H$ of the agent's effective decision distribution is lower-bounded by $H(E)-C_{\text{channel}}$, where $C_{\text{channel}}$ is the capacity consumed by task-relevant content. Formally, let $I(a_t; \mathrm{goal}\mid c_t)$ denote the mutual information between the agent's next action and the correct goal given the current context. A passage retained in $c_t$ is justified if and only if its presence materially increases this quantity, i.e., if knowing it makes the agent meaningfully more likely to take the correct next action. This criterion favors compressed, action-relevant abstractions over raw interaction history: a structured summary of a failed debugging attempt that identifies the root cause and the attempted remediation carries strictly greater mutual information with future correct actions than the full error trace from which it was derived. The principle provides the information-theoretic justification for treating context compression, summarization, and principled forgetting as first-class architectural requirements rather than engineering conveniences, and motivates the three-level memory hierarchy developed in Section~\ref{subsec:d2}.

\subsection{Desiderata of Agent Cybernetics}

The six principles of Section~\ref{sec:agent} are individually necessary but jointly insufficient as engineering guidance as they identify \emph{what} properties a well-designed agent must possess, but not \emph{how} those properties compose into coherent system-level objectives. We synthesize the six principles into three desiderata that organize the research directions of Section~\ref{sec:suggested_research} and the application analyses of Section~\ref{sec:usecases}.

\begin{desideratum}[Reliability]
An agent must perform intended behaviors without causing unauthorized access, cascading failures, or irreversible harm. When it cannot perform a task well, it must fail gracefully and safely. Safety is prioritized over performance for foundation agents.
\end{desideratum}

Reliability draws primarily on Principles~\ref{prin:feedback} (closed-loop feedback), \ref{prin:variety} (requisite variety), and \ref{prin:second_order} (second-order regulation). The reliability challenge for foundation agents is qualitatively distinct from that of classical software systems: failures are not confined to well-typed exceptions or assertion violations, but manifest as \emph{hallucination} (confidently incorrect tool arguments or plans), \emph{feedback blindness} (proceeding with a flawed plan despite clear failure signals in tool outputs), \emph{cascading failure} (early errors propagating silently through long action sequences before detection), and \emph{irreversibility} (actions such as sending emails, executing code, or modifying databases that cannot be undone). The interaction among these failure modes is particularly consequential: a hallucinated plan that is feedback-blind will execute irreversible actions confidently. Reliability engineering for agents must therefore address not failure modes individually but their \emph{composition} and safety priority means that no performance objective justifies proceeding when a reliability boundary has been reached.

\begin{desideratum}[Lifelong Running]
An agent must operate over very long time horizons, potentially indefinitely, processing and leveraging vast accumulated experience. This requires principled memory management: storing, compressing, organizing, and selectively retrieving experience so that past knowledge improves future performance without overwhelming current context.
\end{desideratum}

Lifelong running draws primarily on Principles~\ref{prin:homeostasis} (goal homeostasis and adaptive re-anchoring), \ref{prin:black_box} (black-box environment modelling), and \ref{prin:memory} (context entropy minimization). The central tension is between \emph{retention} and \emph{relevance}: An agent that remembers too little cannot build on past experience, while one that remembers too much buries its own judgment under accumulated noise. Resolving this tension requires not merely a memory system but a \emph{memory architecture} that actively compresses, organizes, and promotes experience across representational levels, which can be addressed by the three-level hierarchy described in Section~\ref{subsec:d2}. Lifelong running also imposes a goal-stability requirement that goes beyond what any single task demands: an agent operating indefinitely must maintain alignment with its objectives across distributional shifts in the environment, changes in task specification, and the accumulation of experience that may bias the agent's effective objective.

\begin{desideratum}[Self-Improvement]
Agents will perform poorly on novel tasks. They must be able to self-improve through structured trial-and-error, i.e., leveraging lifelong running mechanisms for safe exploration and reliability mechanisms to bound that exploration, so that performance improves over time without requiring human intervention.
\end{desideratum}

Self-improvement draws on all six principles, with Principle~\ref{prin:feedback} (closed-loop feedback) providing the core mechanistic loop and Principles~\ref{prin:homeostasis} (goal homeostasis and adaptive re-anchoring), \ref{prin:second_order} (second-order regulation), and \ref{prin:memory} (context entropy minimization) governing the conditions under which self-modification remains safe. The fundamental difficulty is that self-improvement requires the agent to modify the very regulatory structures that ensure its reliability and goal stability: an agent that improves its inference strategy may inadvertently destabilize its existing competencies, and an agent that revises its goal representation in response to experience may drift from its intended objective. The two-level homeostatic architecture of Principle~\ref{prin:homeostasis} (goal homeostasis and adaptive re-anchoring)\footnote{In the rest of this paper, we will discard the explanations of principles for simplicity.} provides the principled resolution: the inner loop constrains self-modification to preserve reliable performance on mastered behaviors, while the outer loop governs controlled excursion into new behavior space, triggering structural revision only when sustained performance evidence establishes that inner-loop adjustment is insufficient. Self-improvement is not the independent capability of foundation agents but constitutively depends on the reliability and lifelong-running implementation.

\section{Suggested Research by Agent Cybernetics}
\label{sec:suggested_research}

The six principles and three desiderata of \textbf{Agent Cybernetics} do not only diagnose existing agent failures, they actively prescribe a research agenda. We organize that agenda by desideratum, identifying the open problems and connecting them to the engineering patterns developed in Section~\ref{sec:usecases}. 

\subsection{Suggested Research by Desideratum 1: Reliability}
\label{subsec:d1}

\textbf{Redundancy and Multi-Agent Systems.}
From Principle~\ref{prin:variety}, a single agent with bounded variety cannot handle all failure modes. Multi-agent designs implement redundancy through variety: the combined system achieves $V(\mathcal{T})$ exceeding any individual component. When individual agents may transiently deviate from their local viability regions (Principle~\ref{prin:homeostasis}), a redundant verifier serves as a collective homeostatic mechanism, detecting individual deviations that the primary agent cannot self-monitor. This mirrors fault-tolerance patterns from classical systems engineering, now applied to LLM-based reasoning. We note that test-time scaling (e.g., majority voting and best-of-$N$)~\citep{zhang2025survey}, multi-agent debate~\citep{du2024improving,zhang2025stop} and LLM-as-a-Judge~\citep{gu2024survey} can be specific implementations of this redundancy mechanism.
Open research questions include how to assign agent roles to maximize collective variety without redundant overlap, how to aggregate conflicting reliability assessments, and how to bound the communication overhead of coordinated verification~\citep{gonzalezpumariega2026reliability,kuntz2025osharm}.

\textbf{World Models for Predictive Reliability.}
A world model $\hat{f}$ satisfying $\hat{o}_{t+1} = \hat{f}(c_t, a_t)$~\citep{schrittwieser2020mastering,hafner2025mastering} enables pre-action simulation: before executing a potentially destructive action, the agent simulates its effects and verifies consistency with desired outcomes. This transforms reliability from a reactive property (i.e., detecting failures after they occur) to a proactive one (i.e., preventing them through predictive simulation). Within the two-level homeostatic architecture of Principle~\ref{prin:homeostasis}, the world model serves dual roles: it implements fast inner-loop verification during routine execution and constitutes the substrate for simulating candidate goal revisions during outer-loop restructuring. Reliability and adaptivity thereby become mutually reinforcing rather than competing objectives. Research directions include learning world models incrementally from agent interaction histories, calibrating confidence in simulated outcomes, and propagating simulation-detected risks into the harness's execution policy.

\textbf{Human-in-the-Loop Approval.}
Principle~\ref{prin:second_order} implies that an agent must detect when it has reached the boundary of its competence and escalate rather than proceed. Human approval gates, i.e., checkpoints at which the agent pauses before executing high-risk actions, are the engineering instantiation of this principle. The harness literature~\citep{lou2026autoharness,zhou2026externalization} increasingly treats such gates as first-class architectural elements. Research challenges include determining which actions warrant escalation without making human oversight a bottleneck, designing approval interfaces that surface relevant context efficiently, and learning escalation thresholds from operator feedback over time.

\subsection{Suggested Research by Desideratum 2: Lifelong Running}
\label{subsec:d2}

\textbf{Three-Level Memory Hierarchy.}
Human cognitive science distinguishes episodic, semantic, and procedural memory~\citep{sumers2023cognitive}. Agent Cybernetics provides an information-theoretic rationale for the memory hierarchy. The progression from episodic to semantic to procedural memory constitutes a sequence of entropy-reduction operations: each level discards instance-specific detail in favor of generalizable and reusable patterns. This unified view connects the evolution from retrieval-augmented generation (RAG)~\citep{lewis2020retrieval} (episodic and semantic memory) to skills~\citep{li2026skillsbench,yang2026autoskill} (procedural memory), and such information-theoretic principle can further guide researchers to design more useful memories beyond current hierarchy, e.g., optimizing the memory with information bottleneck~\citep{tishby2000information,tishby2015deep}.

\textbf{Regular Retraining and On-Policy Distillation.}
Beyond in-context memory, lifelong running ultimately requires model update: periodically integrating accumulated experience into the model weights themselves~\citep{agarwal2024policy,ye2026policy}. The cybernetic framing identifies the correct objective: retraining should reduce the residual variety $V(\mathcal{W}) - V(\mathcal{T})$ by incorporating environmental patterns the current model cannot handle. Continual learning methods that preserve knowledge of prior tasks while acquiring new competencies implement homeostasis at the weight level. Research priorities include designing replay buffers that maximize variety coverage, quantifying forgetting in terms of residual variety, and developing evaluation protocols that distinguish genuine lifelong improvement from noise.

\textbf{World Models as Lifelong Running Substrates.}
A world model serves not only as a reliability instrument (Section~\ref{subsec:d1}) but also as a lifelong running substrate: it compactly represents environment dynamics learned from interaction trajectories, enabling generalization from sparse experiences. The positive-feedback loop between interaction, world model update, and improved action selection constitutes the mechanistic basis for sustained long-term performance improvement.

\subsection{Suggested Research by Desideratum 3: Self-Improvement}
\label{subsec:d3}

\textbf{Trial-and-Error and RL.}
Reinforcement learning is, in essence, a special case of cybernetics: the agent--environment interaction loop, reward signal, and policy update together instantiate the canonical sense-compute-act feedback cycle of classical control theory~\citep{wiener1949cybernetics}. Exploration widens this loop deliberately, wherein actions serve primarily to acquire environmental information rather than accomplish immediate tasks, the black-box principle applied dynamically~\citep{sutton1998reinforcement}. Self-improvement requires an agent to: (1) generate behavioral hypotheses; (2) execute and observe outcomes; (3) update its policy based on prediction error; and (4) consolidate successful behaviors into procedural memory, with the world model generating predictions and providing the substrate for intrinsic reward.

\textbf{Alignment as Second-Order Regularization.}
Alignment seeks to ensure that agent behavior conforms to human intent, and RLHF~\citep{ouyang2022training,rafailov2023direct} has emerged as the predominant method for achieving this in practice. The agent cybernetics reformulates RLHF as second-order regulation, where the reward model $R'$ governs the first-order policy $\pi_\theta$. This reframing has a consequential implication for research prioritization: the accuracy, coverage, and fidelity of $R'$ to the true task objective constitutes the binding constraint on self-improvement capacity. Accordingly, investment in improving $R'$ through principled feedback elicitation, reward model calibration, and coverage-aware data collection is likely to yield greater returns than equivalent effort devoted to scaling $\pi_\theta$ alone.

\textbf{Stability-Adaptivity Trade-off.}
Second-order cybernetics predicts a fundamental tension: self-improvement requires modifying one's own control laws, yet unrestricted self-modification risks destroying previously acquired competencies, the classical \textbf{\emph{stability-adaptivity dilemma}} in continual learning (corresponding to the exploitation-exploration in RL~\citep{sutton1998reinforcement}). The two-level homeostatic architecture of Principle~\ref{prin:homeostasis} resolves this by separating concerns: the \emph{inner loop} preserves reliable performance on mastered behaviors, while the \emph{outer loop} triggers structural revision only when sustained evidence establishes that inner-loop adjustment is insufficient. Multi-agent systems (MASs)~\citep{wooldridge2009introduction} provide a natural implementation of this two-level architecture: individual agents may explore locally while a system-level coordinator ensures the collective viability region $\Omega_{\text{sys}}$ is preserved.

\textbf{Recursive Self-Improvement.}
% Recursive self-improvement, wherein agents improve their own improvement procedures, is the agentic instantiation of von Foerster's second-order cybernetics. The two-level homeostatic architecture applies recursively at each level of this hierarchy, and key open challenges include formalizing convergence criteria for recursive self-modification, bounding revision depth to prevent runaway objective drift, and defining second-order improvement benchmarks that are independent of the first-order task metrics.
Recursive self-improvement, wherein agents improve their own improvement procedures, is the agentic instantiation of von Foerster's second-order cybernetics~\citep{von2003cybernetics}, with the two-level homeostatic architecture applying recursively at each level of the improvement hierarchy~\citep{jiang2025adaptation}. Key open challenges include formalizing convergence criteria for recursive self-modification, bounding revision depth to prevent runaway objective drift~\citep{turner2022parametrically}, and establishing second-order improvement benchmarks that are decoupled from the first-order task metrics~\citep{chollet2019measure}.

\section{Application Domains and Analysis}
\label{sec:usecases}
\input{appendices/use_cases}

\section{Alternative Views}
\label{sec:alternative_views}
\textbf{Engineering is Enough, Science is Not Necessary.} A natural objection is that the history of deep learning demonstrates that scaling (i.e., more data, more parameters, more compute) resolves most problems, rendering principled architectural frameworks unnecessary. We do not dispute the empirical power of scaling and we argue that agent cybernetics addresses a different dimension. Scaling increases the \emph{policy capacity} of the base model: a larger $\pi_\theta$ can represent a richer mapping from context to action. Cybernetics addresses the \emph{control architecture} within which that policy operates: the feedback structure, the goal-maintenance mechanism, the information-channel management. 

\textbf{Agent Cybernetics is Too Philosophical and Not Relevant to Practice.} A second objection holds that cybernetic concepts (e.g., variety, entropy, homeostasis) are too abstract to yield concrete engineering guidance, and that practitioners already know empirically what works. We respectively disagree both claims. First, the six principles yield \emph{specific, falsifiable engineering recommendations}. These are not reformulations of existing practice in philosophical language, but the prescriptions that current harness implementations frequently violate, as evidenced by the failure modes documented in Section~\ref{sec:usecases}. Second, the theoretical framing provides something that empirical trial-and-error cannot: a \emph{common language} for reasoning about failure modes across domains, and a principled basis for predicting which interventions will generalize. 

% Practitioners working on code agents and research agents currently have no formal basis for recognising that FM-CG1 and FM-AR1 are instances of the same failure (feedback-loop opening); cybernetics makes that recognition immediate and transfers the fix accordingly.

\textbf{Agent Cybernetics is Nothing New.} A third objection notes that many of the ideas in Agent Cybernetics have precedents: retrieval-augmented generation resembles episodic memory, chain-of-thought resembles an inner reasoning loop, and RLHF resembles a feedback control system. We agree that these engineering practices are cybernetically coherent, and indeed regard this as evidence for, not against, our position: cybernetics provides the \emph{explanation} for why they work, which is precisely what a scientific foundation should offer. What Agent Cybernetics adds is not the individual components but their \emph{systematic organization and the gaps it exposes}. Three gaps are consistently absent from existing systems regardless of framework. Identifying these gaps and explaining \emph{why} they are gaps in terms of first principles are the contributions of this position paper.

\section{Conclusion}
\label{sec:conclusion}

Foundation agents have outpaced the theoretical frameworks needed to guide their design, leaving the field predominantly engineering-driven. This paper has argued that cybernetics provides the missing scaffold: by mapping six canonical cybernetic laws onto six agent design principles and synthesizing them into three engineering desiderata, Agent Cybernetics reframes reliability, lifelong running, and self-improvement as the desiderata to be engineered from first principles, rather than empirical targets to be approximated through trial and error, and identifies frontier research directions toward this goal. We hope Agent Cybernetics opens a productive research venue and provides the principled foundation that reliable, long-horizon deployment of foundation agents demands.

\bibliography{agent_cybernetics}
\bibliographystyle{plain}

\appendix

\clearpage

\section{Frequent Asked Questions (FAQs)}

\input{appendices/faqs}

\clearpage
\section{Derivation Summary}

\input{appendices/summary}

% \clearpage
% \section{Practitioner Corroboration}
% \label{app:practitioner}
 
% % The Agent Cybernetics framework developed in this paper is independently corroborated by a growing body of practitioner writing on harness engineering and agentic system design. Kemper~\citep{kemper2026cybernetics} draws an explicit parallel between cybernetic control principles and design patterns emerging in the agentic AI engineering community. Related practitioner perspectives have appeared across industry venues~\citep{odysseus2026harness,philschmid2026harness,tonykip2026harness,zhuanlan2026harness}, reinforcing the practical motivation for the theoretical framework presented here.

% \url{https://openai.com/index/harness-engineering/}

% \url{https://www.anthropic.com/engineering/effective-harnesses-for-long-running-agents}

% \url{https://www.philschmid.de/agent-harness-2026}

% \url{https://x.com/odysseus0z/status/2030416758138634583}

% \url{https://sheldonkemper.substack.com/p/cybernetics-for-agentic-ai-designing}

% \url{https://zhuanlan.zhihu.com/p/2010109567162398654}
% % \citep{vaswani2017attention}

% \url{https://x.com/tonykipkemboi/status/2031068470922670471}

% \clearpage

\end{document}

%% file: appendices/use_cases.tex
% \section{Application Domains and Analysis}

We instantiate the framework across three canonical application domains, identifying domain-specific failure modes and engineering patterns derived from the Agent Cybernetics principles.

\subsection{Code Generation}
\label{subsec:code}

Code generation at agent scale, including writing, debugging, refactoring, and extending software over hundreds of reasoning steps, is a feedback control problem in which the plant is the codebase, the regulator is the LLM policy, and the feedback signal is test execution output. Let $c_t$ be the codebase state, $a_t \in \mathcal{A}_{\text{code}}$ an edit action, and $y_t \in \{0,1\}^k$ the binary pass/fail vector of $k$ tests. The closed-loop update is:
\begin{equation}
  c_{t+1} = \text{Edit}(c_t,\, a_t), \qquad
  a_{t+1} \sim \pi_\theta\!\left(\cdot \mid c_t,\, a_t,\, y_t\right).
  \label{eq:code_loop}
\end{equation}
The signal $y_t$ is objective and immediate, making code generation the domain in which closed-loop behavior is most straightforwardly enforceable, however in practice the loop is frequently left open.

\begin{failurebox}{Code Generation: Failure Modes}
\begin{itemize}[leftmargin=*, topsep=0pt, partopsep=0pt, parsep=0pt, itemsep=2pt]

\item \textbf{FM-CG1 (P1): Feedback blindness.} The agent appends test output to context but regenerates the prior plan rather than diagnosing the failure; $a_{t+1}$ is conditioned on $c_t$ but not causally on $y_t$, i.e., the agent fails to response to the feedback from the code execution.
\item \textbf{FM-CG2 (P3): Goal drift.} Over hundreds of edit steps, the effective goal drifts from the original specification toward local consistency, making the current block compile, passing the immediate test. The original API contract is progressively marginalized; inner-loop re-anchoring proves insufficient, requiring outer-loop goal restructuring.
\item \textbf{FM-CG3 (P2): Insufficient action variety.} An agent equipped only with text-editing tools cannot handle build-system manipulation, dependency resolution, or cross-file refactoring, creating a systematic performance floor wherever $V(\mathcal{T}) < V(\mathcal{W})$.

\end{itemize}
\end{failurebox}

\paragraph{Engineering patterns.}
\textbf{P1} requires structurally inescapable feedback. The harness intercepts every edit action, executes the test suite, and re-injects results in a form mandating acknowledgment: \texttt{[test: FAIL] [failing: $T_{\text{fail}}$] [error: $e$] / Identify root cause before proposing fix.} The final instruction is critical: it makes the feedback signal causally upstream of the next action.
\textbf{P3} requires a two-level re-anchoring mechanism. The inner loop re-injects a structured goal state every $k \approx 20$--$50$ steps:
\begin{equation}
  \text{GoalState}_t = \bigl(\text{OriginalSpec},\; \text{CompletedItems},\; \text{PendingItems},\; \text{FailingInvariants}\bigr).
  \label{eq:goal_state}
\end{equation}
When \texttt{FailingInvariants} fails to decrease across $m$ consecutive checkpoints, the outer loop triggers goal reformulation or human escalation rather than continued inner-loop correction.
\textbf{P5} requires pre-commit self-consistency verification: before any non-trivial change, the agent confirms that interface contracts are preserved, call sites are updated, and no unresolved markers appear in modified files.
\textbf{P6} motivates a \emph{skill library}: verified code patterns promoted from raw edit history when their expected mutual information with future correct actions exceeds a threshold $\theta_{\text{skill}}$. Raw history has high entropy but low predictive value, and curated skills can compress history into reusable procedures, which can significantly reduce the entropy for reliable executions.

\begin{recbox}{Code Generation: Engineering Recommendations}
\begin{itemize}[leftmargin=*, topsep=0pt, partopsep=0pt, parsep=0pt, itemsep=2pt]
\item \textbf{R-CG1 (P1):} Require a structured \texttt{[diagnosis]} block (indexing the failing test, tracking root cause, and planning for fix) before any edit following a test failure. Edits without a valid \texttt{[diagnosis]} are rejected by the harness.
\item \textbf{R-CG2 (P3):} At every $k$-th step inject $\text{GoalState}_t$. Classify each requirement as \textsc{done}/\textsc{in-progress}/\textsc{not-started}/\textsc{broken}. If $m$ consecutive checkpoints report the same \textsc{broken} invariants, trigger outer-loop goal reformulation rather than continued inner-loop correction.
\item \textbf{R-CG3 (P5/P6):} Gate all non-trivial commits with a self-consistency check (interface contracts, call-site updates, unresolved markers). Promote resolved novel bug classes to the skill library after validation on at least two held-out instances.
\end{itemize}
\end{recbox}

\begin{table}[ht]
\centering
\caption{Agent cybernetics applied to code generation.}
\label{tab:code}
\small
\begin{tabular}{lll}
\toprule
\textbf{Principle} & \textbf{Domain Instantiation} & \textbf{Engineering Pattern} \\
\midrule
P1 Closed-loop    & Test harness as feedback signal       & Harness-enforced acknowledgment (R-CG1) \\
P3 Goal homeo.    & Two-level spec re-anchoring           & Two-level specification checkpoint (R-CG2) \\
P5 Second-order & Pre-commit consistency check          & Self-consistency gate (R-CG3) \\
P6 Entropy min.   & Skill library as procedural memory    & Verified skill promotion (R-CG3) \\
\bottomrule
\end{tabular}
\end{table}

\subsection{Computer Use}
\label{subsec:computer}

Computer use agents (CUAs) operate by perceiving computer state and issuing actions to manipulate the computer. Two architecturally distinct sub-classes arise naturally from the practices: i) \emph{GUI agents}, which perceive visual screen state through screenshots and act via mouse and keyboard, and ii) \emph{CLI agents}, which perceive structured shell output and act via command invocation. Both instantiate the same abstract control loop, i.e., perceive state, select action, observe effect, but differ fundamentally along three dimensions: feedback signal richness, environment variety, and action reversibility, which will be discussed below in detail.

\paragraph{Control-loop Formulation.}
Let $s_t \in \mathcal{S}$ be the computer state, $a_t \in \mathcal{A}$ an action, and $y_t$ the observed feedback. The shared update structure is:
\begin{equation}
  s_{t+1} = T(s_t, a_t), \qquad
  a_{t+1} \sim \pi_\theta\!\left(\cdot \mid c_t, y_t\right).
  \label{eq:cu_loop}
\end{equation}
For GUI agents, $y_t = \mathrm{Screenshot}(s_{t+1}) \in \mathbb{R}^{H \times W \times 3}$: high-dimensional, pixel-rich, and spatially structured. For CLI agents, $y_t = \mathrm{ShellOutput}(s_{t+1}) \in \Sigma^*$: semantically dense, often machine-parseable, and augmented by exit codes that provide unambiguous success/failure signals analogous to test pass/fail vectors. While GUI feedback is perceptually richer, CLI feedback is epistemically cleaner, but CLI actions are far more consequential and irreversible.

\begin{failurebox}{Computer Use: Failure Modes}
\begin{itemize}[leftmargin=*, topsep=0pt, partopsep=0pt, parsep=0pt, itemsep=2pt]
\item \textbf{FM-GUI1 (P4): Script execution vs.\ exploratory control (GUI).} Agents trained on demonstrations execute predetermined sequences; when the UI changes, they have no recovery behavior because they never built an internal model of UI dynamics. Prior demonstrations are treated as ground truth rather than falsifiable priors.
\item \textbf{FM-GUI2 (P2): Insufficient perceptual variety (GUI).} A click-and-type agent cannot handle drag-and-drop, right-click menus, keyboard shortcuts, or scroll-to-load elements; $V(\mathcal{T}_{\mathrm{GUI}}) < V(\mathcal{W}_{\mathrm{GUI}})$ creates a systematic capability floor.
\item \textbf{FM-GUI3 (P5): Irreversibility blindness (GUI).} The agent treats ``Confirm Delete'' identically to ``Next''; without second-order classification of action reversibility, high-risk actions execute at the same confidence threshold as low-risk ones.
\item \textbf{FM-CLI1 (P1): Exit-code blindness (CLI).} The agent issues a command, observes a non-zero exit code, and proceeds to the next planned command without diagnosis---the CLI analogue of feedback blindness, arising from the same mechanism: local plan continuation over disconfirming signals.
\item \textbf{FM-CLI2 (P2): Insufficient command variety (CLI).} An agent with only basic file-manipulation commands cannot handle package management, process supervision, or network configuration; missing variety in CLI agents may cause dangerous workarounds rather than graceful degradation.
\item \textbf{FM-CLI3 (P5): Catastrophic irreversibility (CLI).} Commands such as \texttt{rm -rf}, \texttt{dd}, or \texttt{DROP TABLE} are immediately and completely irreversible with no system-level confirmation, making the asymmetry between action cost and reversal cost far more extreme than in GUI environments.
\item \textbf{FM-CLI4 (P4): Environment state opacity (CLI).} Background processes, environment variables, file locks, and daemon configurations are not surfaced by default; an agent without a defeasible internal model of accumulated side effects will issue commands whose preconditions are unmet.
\end{itemize}
\end{failurebox}

\paragraph{Engineering Patterns.}
\textbf{P1} requires modality-appropriate feedback gates enforced architecturally. For GUI agents: capture the new screenshot after every atomic action, compare against expected post-action state, and route to a diagnosis branch on mismatch. For CLI agents, the harness enforces explicit exit-code acknowledgment before the next command, with mandatory stderr inspection when $e_t \neq 0$: \texttt{[exit: \$?] [stdout: \ldots] [stderr: \ldots] / If exit $\neq$ 0, diagnose before proceeding.}
\textbf{P2} requires hierarchical action vocabularies matched to environment variety. GUI includes three tiers:  atomic (click, type, scroll), compound (form-fill, dialog-dismiss), and pattern (login flow, pagination). CLI also has three tiers: command, pipeline, script, which are further stratified by privilege level (user, sudoer, root) and scope (local, remote, containerized), since CLI variety includes authorization context, not merely syntactic expressiveness.
\textbf{P3} requires domain-specific goal drift detection. In GUI agents, the inner loop re-anchors before each compound-tier action. In CLI agents, the agent faces \emph{capability-driven objective substitution}, drifting toward tractable sub-goals as principled commands fail, which the outer loop must distinguish from legitimate goal reformulation by requiring explicit justification against the original specification.
\textbf{P4} motivates exploratory probing before consequential actions. For GUI agents: i) visually verify element states, ii) treat action failures as informative observations, and iii) issue low-cost survey actions (hover, scroll) before committing. For CLI agents: i) issue read-only probes (\texttt{ls}, \texttt{env}, \texttt{ps aux}) before write or execution commands, and ii) treat documentation as priors that actual environment behavior may contradict.
\textbf{P5} requires differentiated irreversibility treatment. In GUI agents, application-layer affordances (confirmation dialogs, undo) provide natural second-order regulation cues. In CLI agents, irreversibility is unmediated and demands a strict classification gate: every action $a \in \mathcal{A}_{\mathrm{CLI}}$ carries $r(a) \in \{\mathrm{read}, \mathrm{recoverable}, \mathrm{destructive}\}$, with:
\begin{align}
  \mathrm{Execute}(a_t) \;\;\text{iff}\;\;
  & r(a_t) = \mathrm{read}
  \;\;\vee\;\;
  \bigl( r(a_t) = \mathrm{recoverable} \;\wedge\; \mathrm{Conf}(a_t \mid c_t) > \tau_r \bigr) \notag \\
  & \;\;\vee\;\;
  \bigl( r(a_t) = \mathrm{destructive} \;\wedge\; \mathrm{Conf}(a_t \mid c_t) > \tau_d \;\wedge\; \mathrm{HumanApprove}(a_t) \bigr), \nonumber
\end{align}
where $\tau_d \gg \tau_r$. No confidence threshold alone is sufficient for destructive CLI actions.

\begin{recbox}{Computer Use: Engineering Recommendations}
\begin{itemize}[leftmargin=*, topsep=0pt, partopsep=0pt, parsep=0pt, itemsep=2pt]
\item \textbf{R-CU1 (P1):} \textit{GUI:} bracket every primitive action with \texttt{[expected: X]} / \texttt{[observed: Y]} and mismatch routes to \textsc{Diagnose}. \textit{CLI:} every command is followed by mandatory exit-code acknowledgment and a non-zero exit blocks the next command until a \texttt{[diagnosis]} block is produced.
\item \textbf{R-CU2 (P4):} \textit{GUI:} emit \texttt{[verify: element\_description]} before state-dependent actions and failed verification triggers \textsc{Explore}. \textit{CLI:} issue read-only probes to verify preconditions before any write or execution command and a probe failure triggers an \textsc{Explore} subroutine mapping actual environment state.
\item \textbf{R-CU3 (P5):} \textit{GUI:} require \texttt{[rationale]} before irreversible actions and respect application-layer affordances as safety signals. \textit{CLI:} classify all commands by destructiveness scope and require \texttt{[dry-run]} output and explicit human approval for destructive-scope commands before execution, regardless of model confidence.
\end{itemize}
\end{recbox}

\paragraph{Cross-Modality Observations.} Three cross-cutting patterns emerge from the joint analysis. First, \emph{feedback quality and control conservatism}: CLI feedback is semantically cleaner, but CLI agents should operate \emph{more} conservatively, the higher destructiveness of CLI actions demands a larger safety margin that outweighs the feedback quality advantage. Second, \emph{world model complementarity}: GUI world models represent UI state transitions; CLI world models represent filesystem and process state including side effects. A unified CUA needs both, integrated to maintain a consistent view across modality transitions. Third, \emph{hybrid agent variety}: many real-world tasks require both modalities in sequence and the two-level homeostatic architecture handles this naturally, with the inner loop maintaining goal stability within each modality and the outer loop governing modality-switching decisions when the current modality proves insufficient.

\begin{table}[ht]
\centering
\caption{Agent cybernetics applied to computer use (GUI and CLI).}
\label{tab:cu}
\small
\begin{tabular}{llll}
\toprule
\textbf{Principle} & \textbf{GUI Instantiation} & \textbf{CLI Instantiation} & \textbf{Pattern} \\
\midrule
P1 Closed-loop    & Screenshot comparison gate         & Exit-code acknowledgment gate        & R-CU1 \\
P2 Variety        & Three-tier click/compound/pattern  & Command/pipeline/script + privilege  & ---    \\
P3 Goal homeo.    & UI-context re-anchoring            & Capability-substitution detection    & ---    \\
P4 Black-box      & Pre-action element verification    & Read-only environment probing        & R-CU2 \\
P5 Second-order & Affordance-aware irrev.\ gate      & Dry-run + human approval gate        & R-CU3 \\
\bottomrule
\end{tabular}
\end{table}

\subsection{Automated Research}
\label{subsec:research}

Automated research is the most challenging of the three domains: the agent must search a vast literature, form and revise hypotheses, design and evaluate experiments, and synthesize novel contributions over thousands of reasoning steps. The feedback signal is sparse and high-latency: scientific claims are validated by consistency with a body of evidence accumulated across many interactions, not by immediate test execution.

Let $H_t$ denote the current hypothesis set, $q_t$ a search or experiment action, and $e_t$ the returned evidence. The research loop is:
\begin{equation}
  H_{t+1} = \text{Update}(H_t,\, e_t), \qquad
  q_{t+1} = \text{Design}(H_{t+1},\, \text{ResearchGoal}).
  \label{eq:research_loop}
\end{equation}
Two features distinguish this from the preceding loops: $\text{Update}$ is a belief revision operation rather than a deterministic edit, and the \emph{ResearchGoal} is high-level and highly susceptible to drift.

\begin{failurebox}{Automated Research: Failure Modes}
\begin{itemize}[leftmargin=*, topsep=0pt, partopsep=0pt, parsep=0pt, itemsep=2pt]
\item \textbf{FM-AR1 (P1): Degenerate evidence accumulation.} The agent collects and appends papers but does not perform genuine hypothesis update; each paper is summarized in isolation with no explicit update to $H_t$. The feedback loop of Eq.~\eqref{eq:research_loop} is present in form but $\text{Update}$ is effectively a no-op: $H_{t+1} \approx H_t$ regardless of $e_t$.
\item \textbf{FM-AR2 (P3): Research goal drift.} An agent tasked with studying ``theoretical limits of LLM reasoning'' gradually drifts toward prompt engineering benchmarks, i.e., locally reasonable steps that cumulatively abandon the research question. This is precisely the failure mode outer-loop goal restructuring addresses: inner-loop re-anchoring is insufficient because the diluted goal has become normalized into the working context.
\item \textbf{FM-AR3 (P5): Citation cycling and confirmation bias.} The agent enters a citation loop, repeatedly retrieving and re-summarizing the same cluster of mutually-citing papers, while preferentially selecting evidence that supports the current hypothesis. Both failures are invisible to the agent because it cannot observe its own search behavior patterns.
\end{itemize}
\end{failurebox}

\paragraph{Engineering patterns.}

\textbf{P1} requires hypothesis-driven closed-loop research. After every evidence-gathering action, the agent produces a mandatory structured belief-update block:
\begin{center}
\texttt{[evidence: $e_t$] [supports: $\{h_i : e_t \Rightarrow h_i\}$] [contradicts: $\{h_j : e_t \Rightarrow \neg h_j\}$]}\\
\texttt{[new-hypothesis: $h_{\text{new}}$ if warranted] [next-query: targets highest-uncertainty $h \in H_t$]}
\end{center}
The final line is essential: the next query must be derived from current uncertainty in $H_t$, not from free association, enforcing the closed-loop structure of Eq.~\eqref{eq:research_loop}.
\textbf{P3} requires a monitorable research goal state:
$\text{RGS}_t = \bigl(Q,\; H_t,\; E_t^+,\; E_t^-,\; E_t^?,\; G_t \bigr),$
where $Q$ is the original question, $E_t^{\pm}$ supporting/contradicting evidence, and $G_t \subseteq H_t$ established hypotheses. The inner loop flags drift when $\text{sim}(q_t, Q) < \delta_{\text{drift}}$. The outer loop distinguishes \emph{evidence-driven reformulation} (the agent has discovered $Q$ is ill-posed, supported by accumulated evidence) from \emph{drift-driven degeneration} (the reformulation merely aligns with the current search trajectory).
\textbf{P5} requires a search audit log tracking query diversity, source diversity (breadth of the citation network), and confirmation ratio. When query diversity falls below $\gamma_{\text{diverse}}$ or the confirmation ratio exceeds $\rho_{\text{confirm}}$, the agent must reformulate with new vocabulary, explicitly seek disconfirming evidence, and consider cross-domain analogies.
\textbf{P6} motivates a three-level hierarchical literature memory: (i) \emph{Episodic}: per-paper abstracts and key claims, retrieved by semantic similarity; (ii) \emph{Semantic}: entity-relation knowledge graph derived from many papers; (iii) \emph{Evidential}: a structured evidence table $\{(h_i, e^+_i, e^-_i, \text{conf}_i)\}$ for the active hypothesis set, maintained in-context. The agent operates primarily on Level~3, promoting episodic traces to semantic abstractions when a pattern recurs across $\geq m$ papers, and from semantic to evidential when items are related to open hypotheses.

% \begin{observation}
% The world model in automated research is not an environment dynamics model but a \emph{domain belief state}: a representation of which claims are established, contested, or open. Prediction error---the gap between what the agent expects to find and what it actually finds---is a natural intrinsic reward signal that drives the agent toward information that fills its knowledge gaps.
% \end{observation}

\begin{recbox}{Automated Research: Engineering Recommendations}
\begin{itemize}[leftmargin=*, topsep=0pt, partopsep=0pt, parsep=0pt, itemsep=2pt]
\item \textbf{R-AR1 (P1):} Every evidence-collection step requires a mandatory belief-update step classifying each piece of evidence as supporting, contradicting, or uninformative per active hypothesis. Queries that do not reference the current hypothesis set are rejected.
\item \textbf{R-AR2 (P3):} Maintain $\text{RGS}_t$ as an explicit structured object, re-inject every $k$ steps and flag queries with similarity below $\delta_{\text{drift}}$ to $Q$. If inner-loop re-anchoring fails for $m$ consecutive cycles, trigger a structured goal-reformulation step requiring evidence-based justification and explicit confirmation before adopting a revised $Q'$.
\item \textbf{R-AR3 (P5/P6):} Compute query entropy $H(q_{t-k:t})$ and source entropy $H(s_{t-k:t})$ over a sliding window and inject \textsc{Explore} directives when either drops below threshold. Maintain the Level-3 evidence table as working memory and reconstruct it when P3's outer loop reformulates the research goal.
\end{itemize}
\end{recbox}

\begin{table}[ht]
\centering
\caption{Agent cybernetics applied to automated research.}
\label{tab:res}
\small
\begin{tabular}{lll}
\toprule
\textbf{Principle} & \textbf{Domain Instantiation} & \textbf{Engineering Pattern} \\
\midrule
P1 Closed-loop    & Hypothesis-verify as feedback         & Structured belief update (R-AR1) \\
P3 Goal homeo.    & Two-level RGS monitoring              & Two-level RGS monitor (R-AR2) \\
P5  Second-order & Search diversity and citation cycling & Search diversity monitor (R-AR3) \\
P6 Entropy min.   & Three-level literature memory         & Evidence table as working memory (R-AR3) \\
\bottomrule
\end{tabular}
\end{table}

\subsection{Cross-Domain Analysis}
\label{sec:cross}

% \begin{insightbox}{Central Observations}
% Across all three domains, the most prevalent and consequential failure mode is \emph{feedback blindness}: the feedback signal is structurally present in the agent's context but not causally connected to the subsequent action. The test loop in Eq.~\eqref{eq:code_loop}, the screenshot comparison in the computer use loop, and the belief update in Eq.~\eqref{eq:research_loop} all fail for the same underlying reason: the auto-regressive training objective incentivizes continuation of a locally plausible plan over revision in response to a disconfirming signal. This observation has a universal architectural implication: feedback signals must be structurally enforced, not merely available.
% \end{insightbox}

\begin{table}[ht]
\centering
\caption{Principle salience by application domain (\textbullet{} minor, \textbullet\textbullet{} important, \textbullet\textbullet\textbullet{} critical).}
\label{tab:salience}
\small
\begin{tabular}{lcccc}
\toprule
\textbf{Principles} & \makecell{\textbf{Code}\\\textbf{Generation}} & \makecell{\textbf{Computer}\\\textbf{Use}} & \makecell{\textbf{Auto}\\\textbf{Research}} & \textbf{Reason for Variation} \\

% \textbf{Principle} & \textbf{Code} & \textbf{Computer use} & \textbf{Research} & \textbf{Reason for variation} \\
\midrule
P1 Closed-loop    & $\bullet\bullet\bullet$ & $\bullet\bullet\bullet$ & $\bullet\bullet\bullet$ & Universal; feedback sparsity varies \\
P2 Variety        & $\bullet\bullet$        & $\bullet\bullet\bullet$ & $\bullet$               & UI has highest environmental variety \\
P3 Goal homeo.    & $\bullet\bullet\bullet$ & $\bullet\bullet$        & $\bullet\bullet\bullet$ & Two-level arch.\ active in all domains \\
P4 Black-box      & $\bullet$               & $\bullet\bullet\bullet$ & $\bullet\bullet$        & UI is least documented environment \\
P5 Second-order & $\bullet\bullet$        & $\bullet\bullet\bullet$ & $\bullet\bullet\bullet$ & Irreversibility and loops most dangerous \\
P6 Entropy min.   & $\bullet\bullet$        & $\bullet$               & $\bullet\bullet\bullet$ & Literature search most context-intensive \\
\bottomrule
\end{tabular}
\end{table}

Table~\ref{tab:salience} rates the urgency of each principle across domains. Note that P3 carries elevated salience across all three relative to what single-level treatments would assign: even in computer use, where goal drift has historically been a minor concern, the two-level architecture surfaces a meaningful distinction between inner-loop UI goal maintenance and outer-loop task replanning. The three domains form a gradient of feedback signal quality:
\begin{equation}
  \underbrace{\text{Code generation}}_{\text{binary, immediate}} \succ
  \underbrace{\text{Computer use}}_{\text{rich, immediate}} \succ
  \underbrace{\text{Automated research}}_{\text{sparse, delayed}}. \nonumber
\end{equation}
By the Shannon-Wiener channel capacity principle (P6), lower bandwidth requires more conservative action policies, heavier reliance on internal world models, and more cautious outer-loop triggering since sparse feedback provides less information to distinguish legitimate goal reformulation from inadvertent drift. Of the six principles, P5 (second-order meta-cognitive regulation) is most consistently absent from current systems. P1 is at least partially addressed by modern harness frameworks. P3 is increasingly recognized in long-context agent work, but typically only at the inner-loop level and outer-loop goal restructuring remains largely unimplemented. P5, by contrast, is absent across all three domains: code agents do not detect repetitive edit-fail-edit cycles; computer use agents do not detect fruitless repeated clicks; research agents do not detect collapsed query entropy. All three are instances of the same failure: the absence of a statistical monitor over the agent's own recent action history. Such monitors are low-cost to implement, e.g., statistical functions over the action log requiring no modifications to the underlying model, yet the expected benefit is high, since every failure mode attributable to P5 violations is detectable by simple pattern statistics. We identify P5 meta-cognitive monitoring as the highest-value, lowest-cost intervention  across all three domains.

%% file: appendices/faqs.tex
% ============================================================
% Appendix A: Frequently Asked Questions (FAQs)
% Usage: \input{appendix_a_faq_only}  inside your \appendix
% Requires: \usepackage{xcolor,enumitem,amsmath}
% ============================================================

% \section{Frequent Asked Questions (FAQs)}
% \label{app:faq}

% ── A.1 ─────────────────────────────────────────────────────
\subsection{What Are the Novelties of The Two-level Homeostatic Architecture in Agent Cybernetics?}

The two-level homeostatic architecture derived from Principle~\ref{prin:homeostasis} is grounded in Ashby's ultrastability~\cite{ashby1960design}, whose theoretical antecedents in adaptive systems design are well established.  Our contribution is the transfer of this architecture to the semantic, goal-directed setting of foundation agents, where the ``essential variables'' being regulated are task objectives rather than physiological quantities.  Prior agent frameworks have implicitly implemented aspects of inner-loop re-anchoring but the outer-loop mechanism of \emph{principled revision of the goal representation} when inner-loop corrections persistently fail has not, to our knowledge, been formally articulated as a design requirement.  We regard this as a significant open engineering problem with concrete implementation stakes.

% ── A.2 ─────────────────────────────────────────────────────
\subsection{What Are the Differences between Agent Cybernetics and Engineering Cybernetics?}

Qian's foundational programme, building reliable systems from unreliable modules in uncertain environments~\cite{qian1954engineering}, is not a historical curiosity, which is the engineering problem that modern AI deployment faces precisely.  The LLM is the unreliable module and the agent harness, world model, and homeostatic monitors are the reliability-conferring architecture.  Qian's programme of bringing mathematical rigour to systems engineering is the spirit we seek to instantiate for agent design.

% ── A.3 ─────────────────────────────────────────────────────
% \subsection{What are Negative and Positive Feedback in Agent Systems?}

% In agent systems, both feedback polarities are present.  Negative feedback is implemented by goal monitors, self-consistency checks, and human approval gates.  Positive feedback is implemented by the autoregressive generation process itself, which amplifies initial biases into confidently incorrect plans.  This asymmetry suggests that architectural biases toward \emph{negative} feedback are important: the system's default behavior, absent explicit intervention, should be conservative rather than amplifying.  The two-level homeostatic architecture implements this structurally: the inner loop defaults to conservative goal maintenance, while the outer loop requires sustained evidence before triggering goal revision.

% ── A.4 ─────────────────────────────────────────────────────

% ── A.5 ─────────────────────────────────────────────────────
\subsection{Are the Three Desiderata Independent, or Do They Form a Dependency Hierarchy?}

The three desiderata are not independent objectives that can be pursued in isolation and they form a strict \emph{dependency hierarchy} where each desideratum constitutes a necessary precondition for the next.

\textbf{Reliability} is the foundational layer.  An agent that cannot perform intended behaviors without cascading failure or irreversible harm provides no stable substrate on which cumulative experience can be built.  \textbf{Lifelong Running} presupposes Reliability: only a reliable agent can accumulate experience without compounding errors across long horizons, and only a reliable agent can safely compress, organise, and re-apply that experience without goal drift corrupting the memory structure itself.  \textbf{Self-Improvement} presupposes both: an agent that modifies its own control laws must be reliable enough that self-modification does not destroy prior competencies, and must have lifelong-running mechanisms in place to retain and evaluate the evidence that justifies each modification.

This ordering has a practical implication for research priority. The desiderata should therefore be engineered and evaluated in sequence, with Reliability gating progress toward Lifelong Running, and Lifelong Running gating progress toward Self-Improvement.

% ── A.6 ─────────────────────────────────────────────────────
\subsection{Does Agent Cybernetics Apply Only to LLM-Based Agents, or Broad Agent Systems?}

The six principles of Agent Cybernetics are derived from cybernetic laws whose domain of applicability is not architecturally specific: they characterize any discrete, high-dimensional, self-referential control system that perceives, reasons, and acts across extended horizons.  The LLM-based foundation agent is the motivating instantiation, but the framework does not presuppose it.

% For agents built on symbolic planners, rule-based systems, or classical RL policies, the six principles apply with straightforward reinterpretation.  The Closed-Loop Feedback principle (P1) applies to any agent whose action selection is not conditioned on the results of prior actions.  The Requisite Variety principle (P2) applies to any agent whose state space or action space is smaller than the environment's variety.  Goal Homeostasis (P3) applies to any long-horizon agent whose objective representation can drift under accumulated context.  The Black-Box principle (P4) applies to any agent interacting with opaque external systems.  Second-Order Regulation (P5) applies to any agent capable of monitoring its own inferential process.  Context Entropy Minimization (P6) applies to any agent with a finite-capacity information channel between perception and action.

% The specific engineering instantiations differ across architectures. Feedback blindness in an LLM agent manifests as plan regeneration despite disconfirming tool outputs; in a classical RL agent, the analogous failure is policy updates that do not incorporate reward signals from the most recent trajectory.  The diagnosis differs; the cybernetic category is identical.  Agent Cybernetics is therefore a domain-agnostic theoretical framework whose primary intended application is LLM-based agents, but whose scope extends to any goal-directed agent operating under uncertainty.

% ── A.7 ─────────────────────────────────────────────────────
\subsection{How Does Agent Cybernetics Relate to Multi-Agent Systems?}

The six principles are stated for a single agent, but each generalizes naturally to multi-agent collectives, and several acquire additional analytic content in that setting.

For Requisite Variety (Principle~\ref{prin:variety}), a multi-agent system's collective output variety $V(T_{\mathrm{collective}})$ is not simply the sum of individual agent varieties, which depends critically on coordination architecture. Agents with overlapping action spaces contribute
\emph{redundancy} (reliability) rather than \emph{variety} (coverage).  Effective multi-agent design
therefore requires role assignment that maximizes collective variety without redundant overlap, a non-trivial combinatorial problem.  For Goal Homeostasis (Principle~\ref{prin:homeostasis}), individual agents may transiently deviate from their local viability regions $\Omega_i$ in service of shared system-level objectives; a system-level coordinator must distinguish sanctioned local deviation from inadvertent goal drift, requiring a collective homeostatic mechanism that is irreducible to individual-agent monitors.  For Second-Order Regulation (Principle~\ref{prin:second_order}), multi-agent debate, majority voting, and LLM-as-a-Judge~\cite{gu2024survey} can be interpreted as collective instantiations of second-order regulation: the aggregate system monitors and corrects the outputs of its component agents without modifying the underlying models.

The framework therefore scales to agent collectives by treating the collective system as the regulated entity and applying the six principles at the system level, while recognizing that each principle may require distinct engineering patterns at the collective level compared to the single-agent case. 

\subsection{How Should Agent Cybernetics Principles Be Prioritized in Practice?}

Table~\ref{tab:salience} provides a domain-specific salience analysis, but a domain-agnostic prioritization can be derived from two considerations: (i)~the cost of violation and (ii)~the implementation overhead of compliance.
By the cost criterion, Principle~\ref{prin:feedback} (Closed-Loop Feedback) and Principle~\ref{prin:second_order} (Second-Order Regulation) carry the highest violation cost, which cause feedback blindness and reasoning loops and irreversibility blindness, respectively. Both failure modes are documented across all three application domains in Appendix~\ref{sec:usecases}.
The recommended practical priority ordering is therefore:
\begin{itemize}[leftmargin=*, topsep=0pt, partopsep=0pt, parsep=0pt, itemsep=0pt]
  \item \textbf{P1}, as the structural prerequisite for any feedback-responsive behaviour;
  \item \textbf{P5}, as the highest-value, lowest-cost intervention;
  \item \textbf{P3}, since outer-loop goal restructuring is the most consistently absent mechanism in
current long-horizon agent systems;
  \item \textbf{P6}, since context compression is a prerequisite for sustained long-horizon
operation;
  \item \textbf{P2 and P4}, whose implementation depends on domain-specific tool and environment
analysis.
\end{itemize}
This ordering is a heuristic for bootstrapping compliance, not a claim that lower-priority principles are less theoretically fundamental.

\subsection{Does Agent Cybernetics Have Implications for Agent Safety and Alignment?}

Agent Cybernetics has direct implications for both safety and alignment, though its contributions are
\emph{architectural} rather than value-theoretic.
For \textbf{safety}, Desideratum~1 (Reliability) explicitly priorities safe failure over performance. 
For \textbf{alignment}, Section~\ref{subsec:d3} reformulates RLHF as second-order regulation: the reward model $R'$ governs the first-order policy $\pi_\theta$, and the accuracy and coverage of $R'$ constitutes the binding constraint on alignment quality.  This reframing has a research prioritization implication: investment in improving $R'$ through principled feedback elicitation, reward model calibration, and coverage-aware data collection yields greater expected alignment improvement than equivalent effort devoted to scaling $\pi_\theta$ alone.
More broadly, Agent Cybernetics identifies \emph{goal drift} (Principle~\ref{prin:homeostasis}) as a structural alignment failure mode: an agent whose goal representation drifts over long contexts is behaviorally misaligned regardless of whether its underlying values are correctly specified.  Outer-loop goal restructuring is therefore an alignment mechanism as much as a reliability mechanism, and the two-level homeostatic architecture provides a principled engineering response to the alignment challenge of long-horizon deployment.

\subsection{AI Usage}

AI tools (e.g., Claude and ChatGPT) are used to polish the texts.  We use Nano Banana to generate the icons in the illustrative figures for better visualization and the full figures are organized manually.

%% file: appendices/summary.tex
Table~\ref{tab:derivation} summarizes the one-to-one correspondence between classical cybernetics laws, agentic principles, and their primary design instantiations and Table~\ref{tab:dependency} characterize the dependency between principles and desiderata.

\begin{table}[ht]
\centering
\caption{Derivation chain from classical cybernetics to agent principles.}
\label{tab:derivation}
% \small
\renewcommand{\arraystretch}{1.3}
\begin{tabular}{lll}
\toprule
\textbf{Classical Law} & \textbf{Agentic Principle} & \textbf{Primary Design Instantiation} \\
\midrule
Feedback        & P1: Closed-loop tool use               & Harness engineering \\
Requisite Variety   & P2: Requisite tool variety             & Hierarchical tool organization \\
Homeostasis & P3: Goal homeo.\ \& adaptive re-anch. & Two-level re-anchoring \\
Black-box   & P4: Black-box environment modelling    & Defeasible world model \\
Second-order & P5: Second-order agentic regulation    & Meta-cognitive monitors \\
Channel capacity & P6: Context entropy minimization       & Three-level memory hierarchy \\
\bottomrule
\end{tabular}
\end{table}

\begin{table}[ht]
\centering
\caption{Dependence of Agent Cybernetics Desiderata on Principles.
Primary dependencies (\checkmark) are as identified in Section~3.2;
dashes indicate the principle is not a primary driver.}
\label{tab:dependency}
\renewcommand{\arraystretch}{1.3}
\begin{tabular}{llccc}
\toprule
& & \textbf{D1}  & \textbf{D2} & \textbf{D3} \\
& & Reliability & Lifelong Running & Self-Improvement \\
\hline
\textbf{P1} & Closed-loop feedback    & \checkmark & ---        & \checkmark \\
\textbf{P2} & Requisite variety       & \checkmark & ---        & ---        \\
\textbf{P3} & Goal homeostasis        & ---        & \checkmark & \checkmark \\
\textbf{P4} & Black-box env.\ modelling & ---      & \checkmark & ---        \\
\textbf{P5} & Second-order regulation    & \checkmark & ---        & \checkmark \\
\textbf{P6} & Context entropy min.    & ---        & \checkmark & \checkmark \\
\bottomrule
\end{tabular}
\end{table}

%% file: agent_cybernetics.bib
@article{lou2026autoharness,
  title={{AutoHarness}: improving {LLM} agents by automatically synthesizing a code harness},
  author={Lou, Xinghua and L{\'a}zaro-Gredilla, Miguel and Dedieu, Antoine and Wendelken, Carter and Lehrach, Wolfgang and Murphy, Kevin P},
  journal={arXiv preprint arXiv:2603.03329},
  year={2026}
}

@book{wiener1949cybernetics,
  title={Cybernetics or Control and Communication in the Animal and the Machine},
  author={Wiener, Norbert},
  year={2019},
  publisher={MIT Press}
}

@book{ashby1956introduction,
  title={An Introduction to Cybernetics},
  author={Ashby, William Ross},
  year={1956},
  publisher={Chapman \& Hall}
}

@book{cannon1939wisdom,
  title={The Wisdom of the Body},
  author={Cannon, Walter Bradford},
  year={1939},
  publisher={Norton \& Co.}
}

@book{ashby1960design,
  title={Design for a Brain: The Origin of Adaptive Behaviour},
  author={Ashby, William},
  year={2013},
  publisher={Springer Science \& Business Media}
}

@article{shannon1948mathematical,
  title={A mathematical theory of communication},
  author={Shannon, Claude Elwood},
  journal={The Bell System Technical Journal},
  volume={27},
  number={3},
  pages={379--423},
  year={1948},
  publisher={Nokia Bell Labs}
}

@incollection{von2003cybernetics,
  title={Cybernetics of cybernetics},
  author={Von Foerster, Heinz},
  booktitle={Understanding understanding: Essays on cybernetics and cognition},
  pages={283--286},
  publisher={Springer},
  year = {2003}
}

@article{intelligence2026pi07steerable,
      title={${\pi}_{0.7}$: A Steerable Generalist Robotic Foundation Model with Emergent Capabilities}, 
      author={Physical Intelligence and Bo Ai and Ali Amin and Raichelle Aniceto and Ashwin Balakrishna and Greg Balke and Kevin Black and George Bokinsky and Shihao Cao and Thomas Charbonnier and Vedant Choudhary and Foster Collins and Ken Conley and Grace Connors and James Darpinian and Karan Dhabalia and Maitrayee Dhaka and Jared DiCarlo and Danny Driess and Michael Equi and Adnan Esmail and Yunhao Fang and Chelsea Finn and Catherine Glossop and Thomas Godden and Ivan Goryachev and Lachlan Groom and Haroun Habeeb and Hunter Hancock and Karol Hausman and Gashon Hussein and Victor Hwang and Brian Ichter and Connor Jacobsen and Szymon Jakubczak and Rowan Jen and Tim Jones and Gregg Kammerer and Ben Katz and Liyiming Ke and Mairbek Khadikov and Chandra Kuchi and Marinda Lamb and Devin LeBlanc and Brendon LeCount and Sergey Levine and Xinyu Li and Adrian Li-Bell and Vladislav Lialin and Zhonglin Liang and Wallace Lim and Yao Lu and Enyu Luo and Vishnu Mano and Nandan Marwaha and Aikys Mongush and Liam Murphy and Suraj Nair and Tyler Patterson and Karl Pertsch and Allen Z. Ren and Gavin Schelske and Charvi Sharma and Baifeng Shi and Lucy Xiaoyang Shi and Laura Smith and Jost Tobias Springenberg and Kyle Stachowicz and Will Stoeckle and Jiaming Tang and Jimmy Tanner and Shalom Tekeste and Marcel Torne and Kyle Vedder and Quan Vuong and Anna Walling and Haohuan Wang and Jason Wang and XuDong Wang and Chris Whalen and Samuel Whitmore and Blake Williams and Charles Xu and Sukwon Yoo and Lili Yu and Wuming Zhang and Zhuoyang Zhang and Ury Zhilinsky},
      year={2026},
      journal={arXiv preprint arXiv 2604.15483},
}

@book{wiener1988human,
  title={The Human Use of Human Beings: Cybernetics and Society},
  author={Wiener, Norbert},
  year={1988},
  publisher={Grand Central Publishing}
}

@inproceedings{ouyang2022training,
  title={Training language models to follow instructions with human feedback},
  author={Ouyang, Long and Wu, Jeff and Jiang, Xu and Almeida, Diogo and Wainwright, Carroll L and Mishkin, Pamela and Zhang, Chong and Agarwal, Sandhini and Slama, Katarina and Ray, Alex and others},
  booktitle={NeurIPS},
  pages={27730--27744},
  year={2022}
}

@inproceedings{rafailov2023direct,
  title={Direct preference optimization: your language model is secretly a reward model},
  author={Rafailov, Rafael and Sharma, Archit and Mitchell, Eric and Ermon, Stefano and Manning, Christopher D and Finn, Chelsea},
  booktitle={NeurIPS},
  pages={53728--53741},
  year={2023}
}

@book{qian1954engineering,
  title={Engineering Cybernetics},
  author={Hsue Shen Tsien},
  year={1954},
  publisher={McGraw-Hill, New York}
}

@article{jiang2025adaptation,
  title={Adaptation of agentic {AI}},
  author={Jiang, Pengcheng and Lin, Jiacheng and Shi, Zhiyi and Wang, Zifeng and He, Luxi and Wu, Yichen and Zhong, Ming and Song, Peiyang and Zhang, Qizheng and Wang, Heng and others},
  journal={arXiv preprint arXiv:2512.16301},
  year={2025}
}

@article{sumers2023cognitive,
  title={Cognitive architectures for language agents},
  author={Sumers, Theodore and Yao, Shunyu and Narasimhan, Karthik R and Griffiths, Thomas L},
  journal={Transactions on Machine Learning Research},
  year={2023}
}

@article{gonzalezpumariega2026reliability,
      title={On the Reliability of Computer Use Agents}, 
      author={Gonzalo Gonzalez-Pumariega and Saaket Agashe and Jiachen Yang and Ang Li and Xin Eric Wang},
      year={2026},
      journal={arXiv preprint arXiv:2604.17849},
}

@inproceedings{
kuntz2025osharm,
title={{OS-Harm}: A Benchmark for Measuring Safety of Computer Use Agents},
author={Thomas Kuntz and Agatha Duzan and Hao Zhao and Francesco Croce and J Zico Kolter and Nicolas Flammarion and Maksym Andriushchenko},
booktitle={NeurIPS Datasets and Benchmarks Track},
year={2025},
url={https://openreview.net/forum?id=Di30GwhQSX}
}

@book{sutton1998reinforcement,
  title={Reinforcement Learning: An Introduction},
  author={Sutton, Richard S and Barto, Andrew G},
  year={1998},
  publisher={MIT press Cambridge}
}

@inproceedings{
jimenez2024swebench,
title={{SWE}-bench: Can Language Models Resolve Real-world Github Issues?},
author={Carlos E Jimenez and John Yang and Alexander Wettig and Shunyu Yao and Kexin Pei and Ofir Press and Karthik R Narasimhan},
booktitle={ICLR},
year={2024},
}

@article{deng2025swe,
  title={{SWE}-bench pro: Can {AI} agents solve long-horizon software engineering tasks?},
  author={Deng, Xiang and Da, Jeff and Pan, Edwin and He, Yannis Yiming and Ide, Charles and Garg, Kanak and Lauffer, Niklas and Park, Andrew and Pasari, Nitin and Rane, Chetan and others},
  journal={arXiv preprint arXiv:2509.16941},
  year={2025}
}

@inproceedings{xie2024osworld,
  title={{OSWorld}: benchmarking multimodal agents for open-ended tasks in real computer environments},
  author={Xie, Tianbao and Zhang, Danyang and Chen, Jixuan and Li, Xiaochuan and Zhao, Siheng and Cao, Ruisheng and Hua, Toh Jing and Cheng, Zhoujun and Shin, Dongchan and Lei, Fangyu and others},
  booktitle={NeurIPS},
  pages={52040--52094},
  year={2024}
}

@article{bonatti2024windows,
  title={Windows agent arena: Evaluating multi-modal os agents at scale},
  author={Bonatti, Rogerio and Zhao, Dan and Bonacci, Francesco and Dupont, Dillon and Abdali, Sara and Li, Yinheng and Lu, Yadong and Wagle, Justin and Koishida, Kazuhito and Bucker, Arthur and others},
  journal={arXiv preprint arXiv:2409.08264},
  year={2024}
}

@article{hubert2025olympiad,
  title={Olympiad-level formal mathematical reasoning with reinforcement learning},
  author={Hubert, Thomas and Mehta, Rishi and Sartran, Laurent and Horv{\'a}th, Mikl{\'o}s Z and {\v{Z}}u{\v{z}}i{\'c}, Goran and Wieser, Eric and Huang, Aja and Schrittwieser, Julian and Schroecker, Yannick and Masoom, Hussain and others},
  journal={Nature},
  pages={1--3},
  year={2025},

}

@article{trinh2024solving,
  title={Solving olympiad geometry without human demonstrations},
  author={Trinh, Trieu H and Wu, Yuhuai and Le, Quoc V and He, He and Luong, Thang},
  journal={Nature},
  volume={625},
  number={7995},
  pages={476--482},
  year={2024},
}

@article{feng2026internagent,
  title={Internagent-1.5: A unified agentic framework for long-horizon autonomous scientific discovery},
  author={Feng, Shiyang and Ma, Runmin and Yan, Xiangchao and Fan, Yue and Hu, Yusong and Huang, Songtao and Zhang, Shuaiyu and Cao, Zongsheng and Peng, Tianshuo and Yuan, Jiakang and others},
  journal={arXiv preprint arXiv:2602.08990},
  year={2026}
}

@article{zhang2025multimodal,
  title={A multimodal robotic platform for multi-element electrocatalyst discovery},
  author={Zhang, Zhen and Ren, Zhichu and Hsu, Chia-Wei and Chen, Weibin and Hong, Zhang-Wei and Lee, Chi-Feng and Penn, Aubrey and Xu, Hongbin and Zheng, Daniel J and Miao, Shuhan and others},
  journal={Nature},
  volume={647},
  number={8089},
  pages={390--396},
  year={2025},
}

@article{gao2026autonomous,
  title={Autonomous closed-loop framework for reproducible perovskite solar cells},
  author={Gao, Danpeng and Lu, Shuaihua and Zhang, Chunlei and Wang, Ning and Yu, Zexin and Sun, Xianglang and Martin, Rebecca and Vanin, Francesco and Qian, Liangchen and Long, Nicholas and others},
  journal={Nature},
  pages={1--3},
  year={2026},
  publisher={Nature Publishing Group UK London}
}

@article{qin2024tool,
  title={Tool learning with foundation models},
  author={Qin, Yujia and Hu, Shengding and Lin, Yankai and Chen, Weize and Ding, Ning and Cui, Ganqu and Zeng, Zheni and Zhou, Xuanhe and Huang, Yufei and Xiao, Chaojun and others},
  journal={ACM Computing Surveys},
  volume={57},
  number={4},
  pages={1--40},
  year={2024},
}

@article{xue2025simpletir,
  title={{SimpleTIR}: End-to-end reinforcement learning for multi-turn tool-integrated reasoning},
  author={Xue, Zhenghai and Zheng, Longtao and Liu, Qian and Li, Yingru and Zheng, Xiaosen and Ma, Zejun and An, Bo},
  journal={arXiv preprint arXiv:2509.02479},
  year={2025}
}

@article{chhikara2025mem0,
  title={Mem0: Building production-ready {AI} agents with scalable long-term memory},
  author={Chhikara, Prateek and Khant, Dev and Aryan, Saket and Singh, Taranjeet and Yadav, Deshraj},
  journal={arXiv preprint arXiv:2504.19413},
  year={2025}
}

@article{zhou2026externalization,
  title={Externalization in {LLM} Agents: A Unified Review of Memory, Skills, Protocols and Harness Engineering},
  author={Zhou, Chenyu and Chai, Huacan and Chen, Wenteng and Guo, Zihan and Shan, Rong and Song, Yuanyi and Xu, Tianyi and Yang, Yingxuan and Yu, Aofan and Zhang, Weiming and others},
  journal={arXiv preprint arXiv:2604.08224},
  year={2026}
}

@article{lee2026meta,
  title={Meta-Harness: End-to-End Optimization of Model Harnesses},
  author={Lee, Yoonho and Nair, Roshen and Zhang, Qizheng and Lee, Kangwook and Khattab, Omar and Finn, Chelsea},
  journal={arXiv preprint arXiv:2603.28052},
  year={2026}
}

@inproceedings{shinn2023reflexion,
  title={Reflexion: language agents with verbal reinforcement learning},
  author={Shinn, Noah and Cassano, Federico and Gopinath, Ashwin and Narasimhan, Karthik and Yao, Shunyu},
  booktitle={Proceedings of the 37th International Conference on Neural Information Processing Systems},
  pages={8634--8652},
  year={2023}
}

@inproceedings{
tan2025cradle,
title={Cradle: Empowering Foundation Agents towards General Computer Control},
author={Weihao Tan and Wentao Zhang and Xinrun Xu and Haochong Xia and Ziluo Ding and Boyu Li and Bohan Zhou and Junpeng Yue and Jiechuan Jiang and Yewen Li and Ruyi An and Molei Qin and Chuqiao Zong and Longtao Zheng and YuJie Wu and Xiaoqiang Chai and Yifei Bi and Tianbao Xie and Pengjie Gu and Xiyun Li and Ceyao Zhang and Long Tian and Chaojie Wang and Xinrun Wang and B{\"o}rje F. Karlsson and Bo An and Shuicheng YAN and Zongqing Lu},
booktitle={ICML},
year={2025},
url={https://openreview.net/forum?id=6CAgbrjHTc}
}

@inproceedings{zitkovich2023rt2,
  title={{RT-2}: Vision-language-action models transfer web knowledge to robotic control},
  author={Zitkovich, Brianna and Yu, Tianhe and Xu, Sichun and Xu, Peng and Xiao, Ted and Xia, Fei and Wu, Jialin and Wohlhart, Paul and Welker, Stefan and Wahid, Ayzaan and others},
  booktitle={CoRL},
  pages={2165--2183},
  year={2023},
}

@article{zhang2026hyperagents,
  title={Hyperagents},
  author={Zhang, Jenny and Zhao, Bingchen and Yang, Wannan and Foerster, Jakob and Clune, Jeff and Jiang, Minqi and Devlin, Sam and Shavrina, Tatiana},
  journal={arXiv preprint arXiv:2603.19461},
  year={2026}
}

@article{zhang2025darwin,
  title={{Darwin} {Godel} machine: Open-ended evolution of self-improving agents},
  author={Zhang, Jenny and Hu, Shengran and Lu, Cong and Lange, Robert and Clune, Jeff},
  journal={arXiv preprint arXiv:2505.22954},
  year={2025}
}

@article{zhang2025survey,
  title={A survey on test-time scaling in large language models: What, how, where, and how well?},
  author={Zhang, Qiyuan and Lyu, Fuyuan and Sun, Zexu and Wang, Lei and Zhang, Weixu and Hua, Wenyue and Wu, Haolun and Guo, Zhihan and Wang, Yufei and Muennighoff, Niklas and others},
  journal={arXiv preprint arXiv:2503.24235},
  year={2025}
}

@inproceedings{
du2024improving,
title={Improving Factuality and Reasoning in Language Models through Multiagent Debate},
author={Yilun Du and Shuang Li and Antonio Torralba and Joshua B. Tenenbaum and Igor Mordatch},
booktitle={ICML},
year={2024},
}

@article{zhang2025stop,
  title={Stop Overvaluing Multi-Agent Debate--We Must Rethink Evaluation and Embrace Model Heterogeneity},
  author={Zhang, Hangfan and Cui, Zhiyao and Chen, Jianhao and Wang, Xinrun and Zhang, Qiaosheng and Wang, Zhen and Wu, Dinghao and Hu, Shuyue},
  journal={arXiv preprint arXiv:2502.08788},
  year={2025}
}

@article{gu2024survey,
  title={A survey on {LLM}-as-a-judge},
  author={Gu, Jiawei and Jiang, Xuhui and Shi, Zhichao and Tan, Hexiang and Zhai, Xuehao and Xu, Chengjin and Li, Wei and Shen, Yinghan and Ma, Shengjie and Liu, Honghao and others},
  journal={The Innovation},
  year={2024},
  publisher={Elsevier}
}

@article{schrittwieser2020mastering,
  title={Mastering {Atari}, go, chess and shogi by planning with a learned model},
  author={Schrittwieser, Julian and Antonoglou, Ioannis and Hubert, Thomas and Simonyan, Karen and Sifre, Laurent and Schmitt, Simon and Guez, Arthur and Lockhart, Edward and Hassabis, Demis and Graepel, Thore and others},
  journal={Nature},
  volume={588},
  number={7839},
  pages={604--609},
  year={2020},
}

@article{hafner2025mastering,
  title={Mastering diverse control tasks through world models},
  author={Hafner, Danijar and Pasukonis, Jurgis and Ba, Jimmy and Lillicrap, Timothy},
  journal={Nature},
  volume={640},
  number={8059},
  pages={647--653},
  year={2025},
}

@inproceedings{lewis2020retrieval,
  title={Retrieval-augmented generation for knowledge-intensive {NLP} tasks},
  author={Lewis, Patrick and Perez, Ethan and Piktus, Aleksandra and Petroni, Fabio and Karpukhin, Vladimir and Goyal, Naman and K{\"u}ttler, Heinrich and Lewis, Mike and Yih, Wen-tau and Rockt{\"a}schel, Tim and others},
  booktitle={NeurIPS},
  pages={9459--9474},
  year={2020}
}

@article{li2026skillsbench,
  title={{SkillsBench}: Benchmarking how well agent skills work across diverse tasks},
  author={Li, Xiangyi and Chen, Wenbo and Liu, Yimin and Zheng, Shenghan and Chen, Xiaokun and He, Yifeng and Li, Yubo and You, Bingran and Shen, Haotian and Sun, Jiankai and others},
  journal={arXiv preprint arXiv:2602.12670},
  year={2026}
}

@article{tishby2000information,
  title={The information bottleneck method},
  author={Tishby, Naftali and Pereira, Fernando C and Bialek, William},
  journal={arXiv preprint physics/0004057},
  year={2000}
}

@inproceedings{tishby2015deep,
  title={Deep learning and the information bottleneck principle},
  author={Tishby, Naftali and Zaslavsky, Noga},
  booktitle={2015 ieee information theory workshop (itw)},
  pages={1--5},
  year={2015},
  organization={Ieee}
}

@inproceedings{agarwal2024policy,
  title={On-policy distillation of language models: Learning from self-generated mistakes},
  author={Agarwal, Rishabh and Vieillard, Nino and Zhou, Yongchao and Stanczyk, Piotr and Garea, Sabela Ramos and Geist, Matthieu and Bachem, Olivier},
  booktitle={ICLR},
  year={2024}
}

@article{ye2026policy,
  title={On-policy context distillation for language models},
  author={Ye, Tianzhu and Dong, Li and Wu, Xun and Huang, Shaohan and Wei, Furu},
  journal={arXiv preprint arXiv:2602.12275},
  year={2026}
}

@book{wooldridge2009introduction,
  title={An Introduction to Multiagent Systems},
  author={Wooldridge, Michael},
  year={2009},
  publisher={John wiley \& sons}
}

@inproceedings{turner2022parametrically,
  title={Parametrically retargetable decision-makers tend to seek power},
  author={Turner, Alexander Matt and Tadepalli, Prasad},
  booktitle={NeurIPS},
  pages={31391--31401},
  year={2022}
}

@article{chollet2019measure,
  title={On the measure of intelligence},
  author={Chollet, Fran{\c{c}}ois},
  journal={arXiv preprint arXiv:1911.01547},
  year={2019}
}

@article{yang2026autoskill,
  title={Autoskill: Experience-driven lifelong learning via skill self-evolution},
  author={Yang, Yutao and Li, Junsong and Pan, Qianjun and Zhan, Bihao and Cai, Yuxuan and Du, Lin and Zhou, Jie and Chen, Kai and Chen, Qin and Li, Xin and others},
  journal={arXiv preprint arXiv:2603.01145},
  year={2026}
}
